\documentclass[11pt]{article}

\usepackage[a4paper, margin=1in]{geometry}
\usepackage{times}
\usepackage{latexsym}
\usepackage[T1]{fontenc}
\usepackage[utf8]{inputenc}
\usepackage{microtype}
\usepackage{inconsolata}
\usepackage{graphicx}
\usepackage{float}
\usepackage{xcolor}
\usepackage{tikz}
\usepackage{booktabs}
\usepackage{array}
\usepackage{tabularx}
\usepackage{enumitem}
\usepackage{amsmath}
\usepackage{ragged2e}
\usepackage[most]{tcolorbox}
\usepackage{url}
\usepackage{seqsplit}
\usepackage[round,authoryear]{natbib}  % provides \citep / \citet
\usepackage{hyperref}
\hypersetup{colorlinks=true, linkcolor=blue!50!black, citecolor=blue!50!black, urlcolor=blue!50!black}

\newcolumntype{Y}{>{\RaggedRight\arraybackslash}X}
\newcolumntype{L}[1]{>{\RaggedRight\arraybackslash}p{#1}}
\newcolumntype{C}[1]{>{\centering\arraybackslash}p{#1}}

\definecolor{plnavy}{HTML}{23395B}
\definecolor{plgray}{HTML}{F3F5F8}

\title{Parthenon Law: A Self-Evolving Legal-Agent Framework}

\author{
  Hejia Geng \\
  \texttt{hejia@tapntell.ai}
  \and
  Leo Liu \\
  \texttt{leo.liu@tapntell.ai}
}

\begin{document}
\maketitle

\begin{abstract}
As agents grow more capable, legal-domain LLM agents promise to turn
document-heavy matters into reviewable work products---yet reliable
deployment faces three obstacles: no large-scale evidence on how today's
strongest model-and-harness combinations behave on end-to-end legal
matters; no agent architecture adapted to the legal vertical, only
general-purpose harnesses; and, in a setting that keeps shifting with new
facts, authorities, and deadlines, no mechanism for systems to learn from
their own outcomes. We address each. A large-scale empirical study on
Harvey LAB---$12{,}510$ agent trajectories---shows that even frontier
agents remain far from completing matters in a single pass: per-criterion
accuracy climbs with stronger models while strict matter completion
stalls. We then introduce \textsc{Parthenon}, a self-evolving
legal-agent framework that factors Model, Harness, Agent roles, legal
Knowledge, deterministic Tools, and procedural Skills into auditable
surfaces for source traceability, date and number grounding, deliverable
compliance, and issue closure. Finally, an anti-leakage learning
loop converts scored failures into task-agnostic edits to skills, tools,
and knowledge, letting the system improve with experience---as a firm
refines its checklists and playbooks after each matter---without touching
model weights. Across our large-scale empirical analysis, \textsc{Parthenon}
substantially improves the performance of state-of-the-art models and
harnesses on legal-matter tasks.
\end{abstract}

\begin{center}\small
\raisebox{-0.2\height}{\includegraphics[height=1.05em]{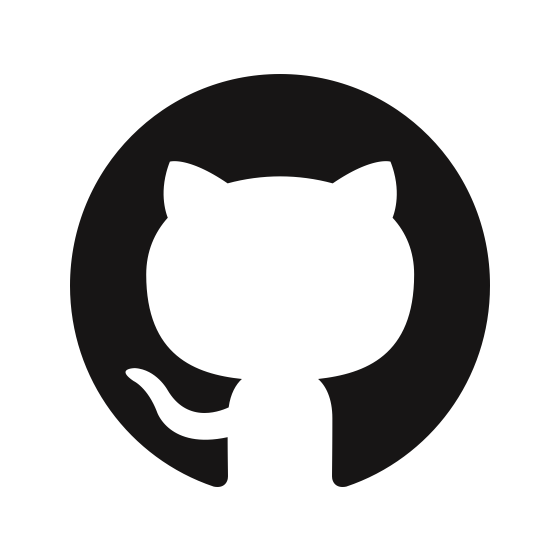}}~%
Code \& skills: \href{https://github.com/HHHHHejia/parthenon-skills}{\texttt{github.com/HHHHHejia/parthenon-skills}}\\[4pt]
\raisebox{-0.2\height}{\includegraphics[height=1.05em]{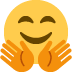}}~%
Data, trajectories \& artifacts: \href{https://huggingface.co/buckets/hhhhhhejia/parthenon-harvey-results-backup}{Hugging Face results backup}
\end{center}

\section{Introduction}
\label{sec:intro}

The field is moving from assistant-style drafting toward agents that
ingest a full matter record, reason across facts and law, and return a
reviewable work product. Errors in this setting are consequential:
missed deadlines, unsupported calculations, omitted issues, invalid
deliverables, or advice untraceable to the record.
Legal benchmarks evaluate reasoning, retrieval, and factuality
\citep{guha2023legalbench,pipitone2024legalbenchrag,
li2024legalagentbench}; Harvey's Legal Agent Benchmark (LAB) extends
this to long-horizon matter work with source documents, deliverables,
and expert rubrics \citep{harvey2026lab}. Yet we still lack a
large-scale empirical analysis of how today's strongest models and
harnesses actually perform on end-to-end legal matters, a framework that
protects legal invariants, and a safe mechanism for turning scored
failures into reusable improvements. These gaps hinder the deployment of
legal AI agents in real-world legal practice.

To measure this gap, we evaluate the major families of legal-agent
execution on the full Harvey LAB corpus: direct API prompting,
a basic legal-native harness, and the Codex and Claude Code workspace
harnesses, each run across model tiers.
LAB's all-criteria-must-pass grading resembles human legal review: a
memo that misses a material deadline, authority, risk, or requested
deliverable remains incomplete. Under this standard, per-criterion scores improve with stronger
models, but strict matter completion stays low. The same failure
modes recur across harnesses and model upgrades: incomplete source
coverage, lost quantitative detail, malformed work products,
unfinished issue analysis, and weak grounding. The bottleneck is
not the model alone; it is the absence of a structured legal work
system around it.

\begin{figure}[H]
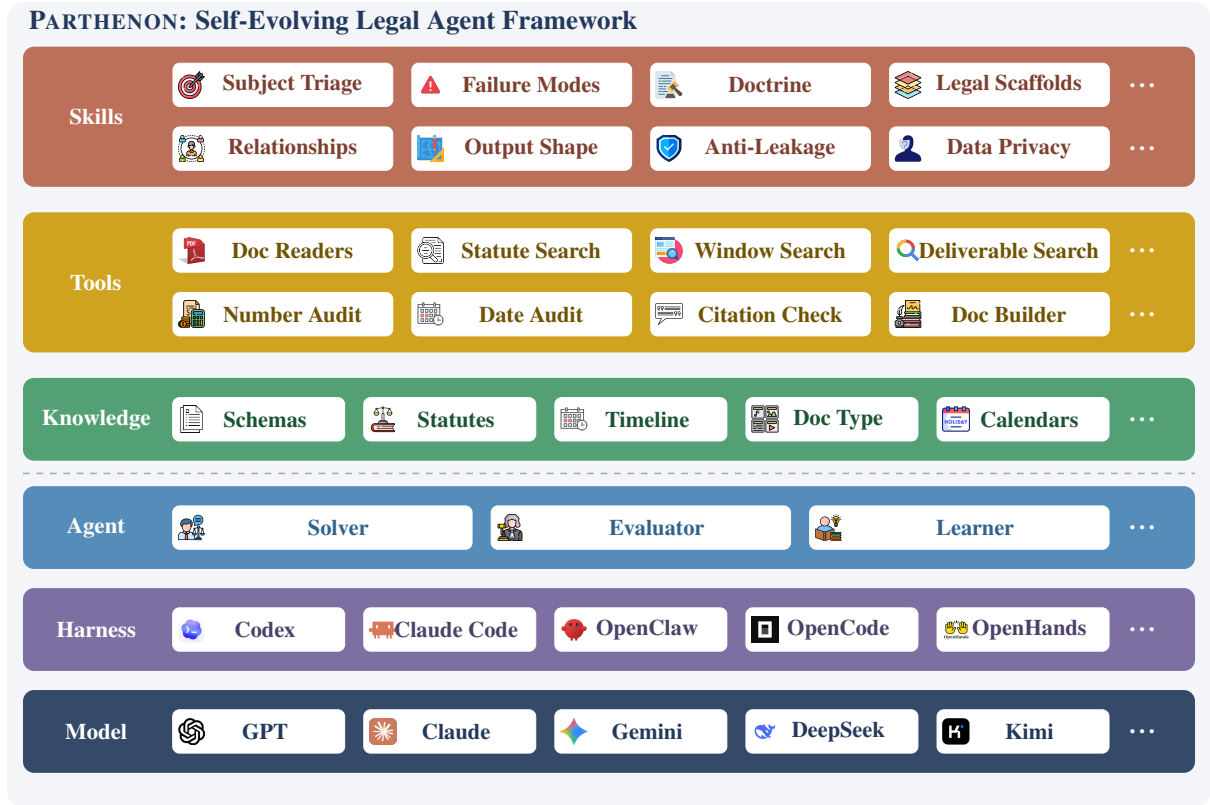

    \centering
    \resizebox{\textwidth}{!}{%
    \begin{tikzpicture}[
      x=1cm, y=1cm,
      layerbox/.style={
        rounded corners=5pt,
        inner sep=0pt
      },
      layerlabel/.style={
        anchor=center,
        text=white,
        font=\bfseries\small,
        align=center,
        text width=2.0cm
      },
      logoblock/.style={
        rounded corners=3pt,
        fill=white,
        draw=white,
        minimum height=0.68cm,
        minimum width=2.70cm,
        inner sep=0pt,
      },
      toolblock/.style={
        rounded corners=3pt,
        fill=white,
        draw=white,
        minimum height=0.68cm,
        minimum width=3.45cm,
        inner sep=0pt,
      },
      agentlogoblock/.style={
        rounded corners=3pt,
        fill=white,
        draw=white,
        minimum height=0.68cm,
        minimum width=4.70cm,
        inner sep=0pt,
      },
    ]
    \definecolor{plnavy}{HTML}{23395B}
    \definecolor{plblue}{HTML}{3E7CB1}
    \definecolor{plpurple}{HTML}{6B5B95}
    \definecolor{plgreen}{HTML}{2E8B57}
    \definecolor{plgold}{HTML}{C99700}
    \definecolor{plred}{HTML}{B35C44}
    \definecolor{plgray}{HTML}{F3F5F8}
    \definecolor{plline}{HTML}{6B7280}
    
    % Outer box
    \fill[rounded corners=8pt, fill=plgray]
      (-0.15,0.55) rectangle (18.75,13.20);
    
    \node[anchor=west, text=plnavy, font=\bfseries\large]
      at (0.05,12.88)
      {\textsc{Parthenon}: Self-Evolving Legal Agent Framework};
    
    % ============ SKILLS LAYER ============
    \fill[layerbox, fill=plred!88] (0.1,10.30) rectangle (18.5,12.50);
    \node[layerlabel] at (1.25,11.40) {Skills};
    \foreach \xpos/\logo/\name in {
      4.18/target/{Subject Triage},
      7.93/warning/{Failure Modes},
      11.68/gavel/Doctrine,
      15.43/layers/{Legal Scaffolds}}{
      \node[toolblock] at (\xpos,11.90) {};
      \node[anchor=west, inner sep=0pt] at (\xpos-1.65,11.90)
        {\includegraphics[height=0.42cm, keepaspectratio]{figures/logos/skills/\logo.png}};
      \node[anchor=center, font=\small\bfseries, text=plred!70!black]
        at (\xpos+0.15,11.90) {\name};
    }
    \foreach \xpos/\logo/\name in {
      4.18/relationship/Relationships,
      7.93/blueprint/{Output Shape},
      11.68/shield/{Anti-Leakage},
      15.43/anonymity/{Data Privacy}}{
      \node[toolblock] at (\xpos,10.90) {};
      \node[anchor=west, inner sep=0pt] at (\xpos-1.65,10.90)
        {\includegraphics[height=0.42cm, keepaspectratio]{figures/logos/skills/\logo.png}};
      \node[anchor=center, font=\small\bfseries, text=plred!70!black]
        at (\xpos+0.15,10.90) {\name};
    }
    \node[font=\bfseries\normalsize, text=white] at (17.70,11.90) {\ldots};
    \node[font=\bfseries\normalsize, text=white] at (17.70,10.90) {\ldots};
    
    % ============ TOOLS LAYER ============
    \fill[layerbox, fill=plgold!88] (0.1,7.70) rectangle (18.5,9.90);
    \node[layerlabel] at (1.25,8.80) {Tools};
    \foreach \xpos/\logo/\name in {
      4.18/reader/{Doc Readers},
      7.93/search/{Statute Search},
      11.68/search2/{Window Search},
      15.43/search3/{Deliverable Search}}{
      \node[toolblock] at (\xpos,9.30) {};
      \node[anchor=west, inner sep=0pt] at (\xpos-1.65,9.30)
        {\includegraphics[height=0.42cm, keepaspectratio]{figures/logos/tools/\logo.png}};
      \node[anchor=center, font=\small\bfseries, text=plgold!55!black]
        at (\xpos+0.15,9.30) {\name};
    }
    \foreach \xpos/\logo/\name in {
      4.18/number/{Number Audit},
      7.93/date/{Date Audit},
      11.68/citation/{Citation Check},
      15.43/docbuilder/{Doc Builder}}{
      \node[toolblock] at (\xpos,8.30) {};
      \node[anchor=west, inner sep=0pt] at (\xpos-1.65,8.30)
        {\includegraphics[height=0.42cm, keepaspectratio]{figures/logos/tools/\logo.png}};
      \node[anchor=center, font=\small\bfseries, text=plgold!55!black]
        at (\xpos+0.15,8.30) {\name};
    }
    \node[font=\bfseries\normalsize, text=white] at (17.70,9.30) {\ldots};
    \node[font=\bfseries\normalsize, text=white] at (17.70,8.30) {\ldots};
    
    % ============ KNOWLEDGE LAYER ============
    \fill[layerbox, fill=plgreen!82] (0.1,6.00) rectangle (18.5,7.30);
    \node[layerlabel] at (1.25,6.65) {Knowledge};
    \foreach \xpos/\logo/\name in {
      3.80/document/Schemas,
      6.80/statute/Statutes,
      9.80/timeline/Timeline,
      12.80/doctype/{Doc Type},
      15.80/holiday/Calendars}{
      \node[logoblock] at (\xpos,6.65) {};
      \node[anchor=west, inner sep=0pt] at (\xpos-1.27,6.65)
        {\includegraphics[height=0.42cm, keepaspectratio]{figures/logos/knowledge/\logo.png}};
      \node[anchor=center, font=\small\bfseries, text=plgreen!60!black]
        at (\xpos+0.10,6.65) {\name};
    }
    \node[font=\bfseries\normalsize, text=white] at (17.70,6.65) {\ldots};
    
    \draw[plline!60, dashed, line width=0.7pt] (0.1,5.80) -- (18.5,5.80);
    
    % ============ AGENT LAYER ============
    \fill[layerbox, fill=plblue!88] (0.1,4.30) rectangle (18.5,5.60);
    \node[layerlabel] at (1.25,4.95) {Agent};
    \foreach \xpos/\logo/\name in {
      4.80/solver/Solver,
      9.80/judge/Evaluator,
      14.80/learn/Learner}{
      \node[agentlogoblock] at (\xpos,4.95) {};
      \node[anchor=west, inner sep=0pt] at (\xpos-2.27,4.95)
        {\includegraphics[height=0.42cm, keepaspectratio]{figures/logos/agent/\logo.png}};
      \node[anchor=center, font=\small\bfseries, text=plblue!80!black]
        at (\xpos+0.25,4.95) {\name};
    }
    \node[font=\bfseries\normalsize, text=white] at (17.70,4.95) {\ldots};
    
    % ============ HARNESS LAYER ============
    \fill[layerbox, fill=plpurple!88] (0.1,2.70) rectangle (18.5,4.00);
    \node[layerlabel] at (1.25,3.35) {Harness};
    \foreach \xpos/\logo/\name in {
      3.80/codex/Codex,
      6.80/claudecode/{Claude Code},
      9.80/openclaw/OpenClaw,
      12.80/opencode/OpenCode,
      15.80/openhands/OpenHands}{
      \node[logoblock] at (\xpos,3.35) {};
      \node[anchor=west, inner sep=0pt] at (\xpos-1.27,3.35)
        {\includegraphics[height=0.42cm, keepaspectratio]{figures/logos/\logo.png}};
      \node[anchor=center, font=\small\bfseries, text=plpurple!70!black]
        at (\xpos+0.10,3.35) {\name};
    }
    \node[font=\bfseries\normalsize, text=white] at (17.70,3.35) {\ldots};
    
    % ============ MODEL LAYER ============
    \fill[layerbox, fill=plnavy!92] (0.1,1.10) rectangle (18.5,2.40);
    \node[layerlabel] at (1.25,1.75) {Model};
    \foreach \xpos/\logo/\name in {
      3.80/openai/GPT,
      6.80/claude/Claude,
      9.80/gemini/Gemini,
      12.80/deepseek/DeepSeek,
      15.80/kimi/Kimi}{
      \node[logoblock] at (\xpos,1.75) {};
      \node[anchor=west, inner sep=0pt] at (\xpos-1.27,1.75)
        {\includegraphics[height=0.42cm, keepaspectratio]{figures/logos/\logo.png}};
      \node[anchor=center, font=\small\bfseries, text=plnavy]
        at (\xpos+0.10,1.75) {\name};
    }
    \node[font=\bfseries\normalsize, text=white] at (17.70,1.75) {\ldots};
    
    \end{tikzpicture}%
    }
    
    \caption{\textbf{\textsc{Parthenon} as a six-layer legal-agent framework.}
    Each layer is a replaceable surface. The Model and Harness layers
    enumerate compatible model families and workspace runtimes; the Agent
    layer fixes the solver--evaluator--learner roles; the Knowledge and
    Tools layers hold the legal knowledge bases and the auditable
    inspection, retrieval, computation, grounding, and validation
    operations over them; and the Skills layer holds the procedural
    scaffold edited by the learner.}
    \label{fig:parthenon-tech-stack}
\end{figure}

\textsc{Parthenon} addresses these gaps as a six-layer framework
organized around attribution and auditability
(Figure~\ref{fig:parthenon-tech-stack}). The lower layers make
execution explicit: Model (LLM capability), Harness (the workspace
runtime), and Agent roles. The upper layers hold legal expertise in
editable artifacts: Knowledge for matter state, concepts, calendars,
and schemas; Tools for deterministic inspection, search,
transformation, and post-draft audits; and Skills for work plans and
release checks. Separating these layers keeps model choice, legal
memory, tool behavior, and procedural guidance from collapsing into
one opaque prompt, while making each failure assignable to a surface
that can actually be edited.

Each scored matter is feedback on the upper harness rather than
data the model must memorize: a solver produces the work product, a
rubric-isolated evaluator converts it into redacted feedback, and a
learner proposes task-agnostic edits to skills, tools, or knowledge.
This mirrors how legal practice firms update checklists, forms, and
playbooks after poor outcomes. An anti-leakage protocol keeps rubric
text, task identifiers, source facts, and answer keys out of the
learner, so candidate edits must generalize beyond the batch that
exposed the failure. The system therefore adapts without fine-tuning
model weights or memorizing benchmark signal.

We run $12{,}510$ agent trajectories on Harvey LAB across the Codex and
Claude Code harnesses and three model tiers. Current harnesses fall
short: a stronger base model lifts per-criterion accuracy but rarely
passes a matter in full, and even the strongest baseline passes every criterion
on barely one matter in eight. With the model and agent harness fixed,
adding \textsc{Parthenon} raises pooled accuracy by
$+13.8/+10.2/+7.4$\,pp---to $82.0/89.9/90.2\%$, a gain comparable to a
model upgrade, from the harness alone---and roughly triples strict all-pass
completion on the weaker solvers ($14\to42$, $47\to137$).

In summary, our contributions are: (i)~a large-scale empirical analysis
of major agent execution modes on Harvey LAB; (ii)~\textsc{Parthenon}, a
six-layer legal-agent framework with auditable surfaces;
and (iii)~a self-evolving learning loop that turns scored failures into
task-agnostic harness updates.

\begin{tcolorbox}[
  breakable, enhanced,
  colback=plgray, colframe=plnavy, boxrule=0.8pt, arc=3pt,
  left=9pt, right=9pt, top=5pt, bottom=6pt,
  coltitle=white, colbacktitle=plnavy, fonttitle=\bfseries,
  title=Key Takeaways]
\setlength{\parskip}{0pt}
\begin{itemize}[leftmargin=1.25em, itemsep=3.5pt, topsep=2pt, parsep=0pt]
  \item \textbf{Procedural knowledge lifts even the strongest models.}
  Adding task-specific skills to a state-of-the-art solver delivers
  gains comparable to a full model upgrade.

  \item \textbf{Failures are mechanical and shared.} Five error
  classes---missing facts, numbers and dates, legal-rule use, deliverable
  form, and coverage---carry roughly two thirds of all failures, and their
  mix is stable across all $24$ practice areas.

  \item \textbf{The hard deficits are model-bound.} Agents lack a notion of
  joint completion, reliable recall of facts and figures from long inputs
  (the largest and fastest-growing error classes), and stable matter
  identity; more reasoning budget is non-monotone and can entrench a wrong
  approach rather than fix these.

  \item \textbf{The workflow deficits are mechanism-bound.} Baselines draft
  in a single pass with no release gate---claiming coverage of sources they
  never open, dropping required sections and shape mid-edit, and citing
  authorities that need not resolve to the record---failures an auditable
  harness can close.

  \item \textbf{An auditable harness buys a model-upgrade-sized lift for
  free.} With the model and runtime fixed, \textsc{Parthenon} adds
  $+13.8/+10.2/+7.4$\,pp (to $82.0/89.9/90.2\%$), largest where the
  baseline is weakest, and on GPT-5.5 it raises accuracy while
  \emph{lowering} per-matter cost.

  \item \textbf{The mechanism is audited procedure, not more text or
  compute.} The gain is a new tool/script bucket; output length is
  decoupled from accuracy, raw reasoning budget is non-monotone, and a
  cached summary does not help---yet a gated learning loop drives two
  unrelated base models to within $0.4$\,pp.

  \item \textbf{Not an autonomous lawyer, but already useful.} The system
  clears every criterion on only a minority of matters, yet runs at
  roughly $70\times$ the speed and thousands of times lower cost---shifting
  the lawyer from drafting to reviewing a source-grounded, audit-flagged
  draft.
\end{itemize}
\end{tcolorbox}

% ---------------------------------------------------------------------------
% arXiv / Overleaf dependency anchors.
% Figure~\ref{fig:parthenon-tech-stack} above builds its icon paths with a
% \foreach \logo loop variable. The static submission scanners (arXiv AutoTeX,
% Overleaf's "Submit to arXiv" export) cannot expand \logo, so they treat every
% loop-only image as an unused orphan, drop it, and the build then fails with
% "File `figures/logos/.../\logo.png' not found". Referencing each file with a
% literal path inside a discarded box forces both the scanners and pdflatex's
% -recorder pass to retain them. This box is assigned but never typeset, so it
% adds nothing to the page.
% ---------------------------------------------------------------------------
\setbox0=\hbox{%
  \includegraphics[height=1pt]{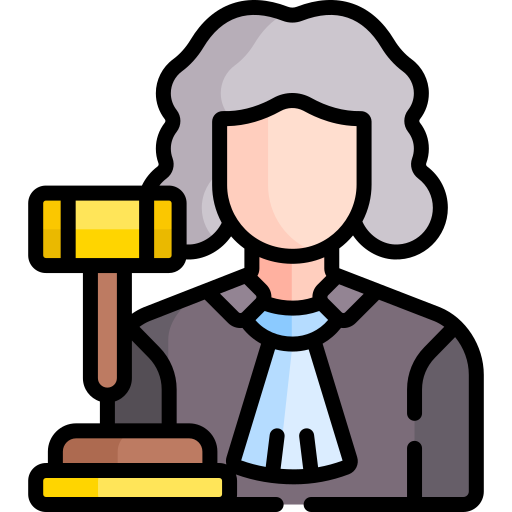}%
  \includegraphics[height=1pt]{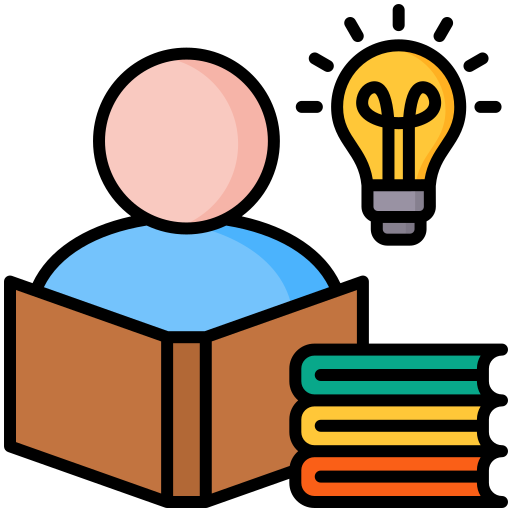}%
  \includegraphics[height=1pt]{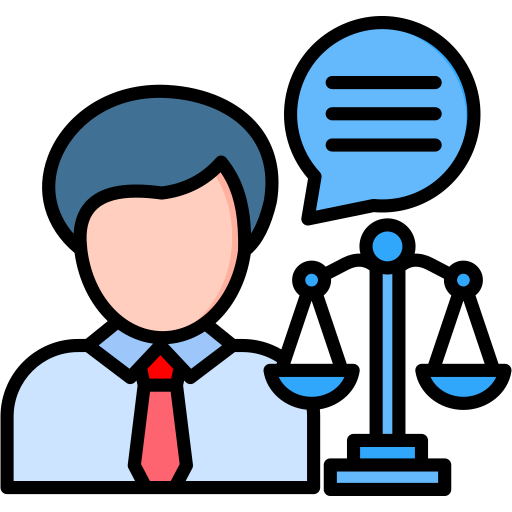}%
  \includegraphics[height=1pt]{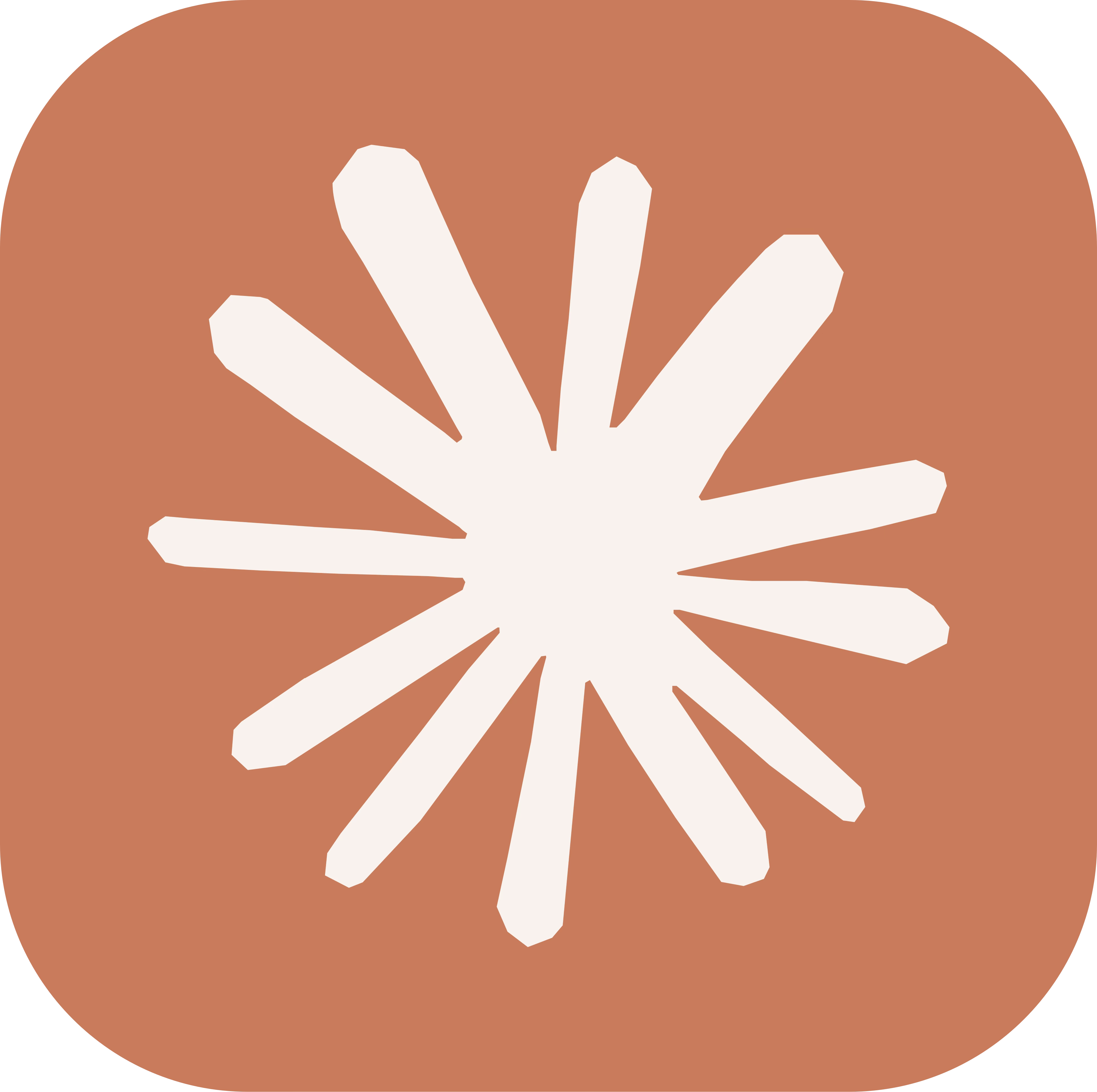}%
  \includegraphics[height=1pt]{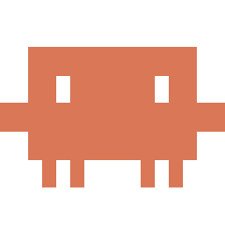}%
  \includegraphics[height=1pt]{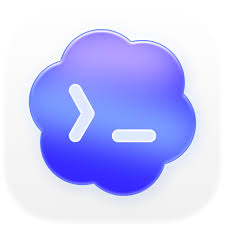}%
  \includegraphics[height=1pt]{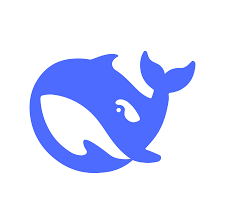}%
  \includegraphics[height=1pt]{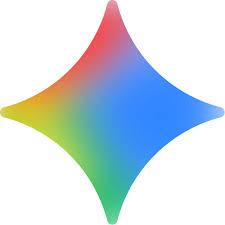}%
  \includegraphics[height=1pt]{figures/logos/github.png}%
  \includegraphics[height=1pt]{figures/logos/huggingface.png}%
  \includegraphics[height=1pt]{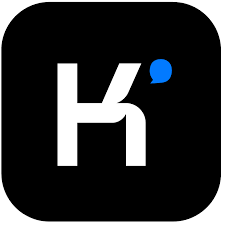}%
  \includegraphics[height=1pt]{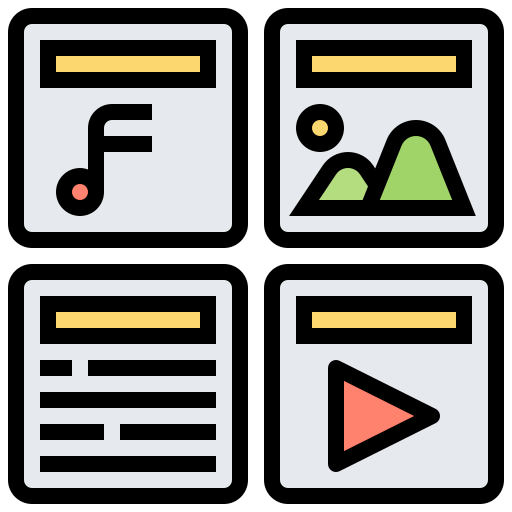}%
  \includegraphics[height=1pt]{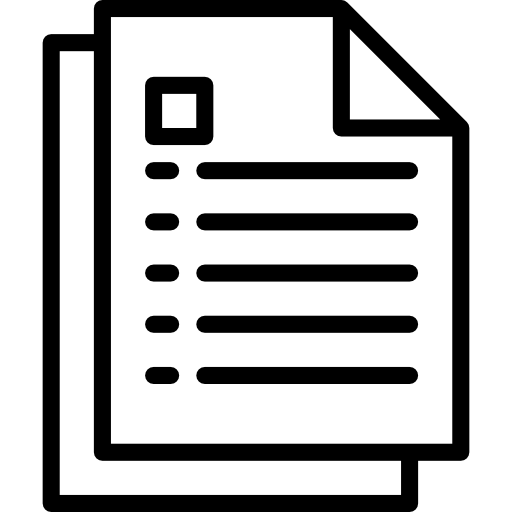}%
  \includegraphics[height=1pt]{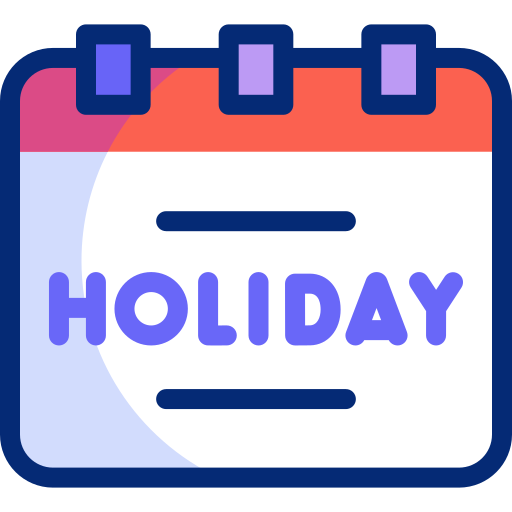}%
  \includegraphics[height=1pt]{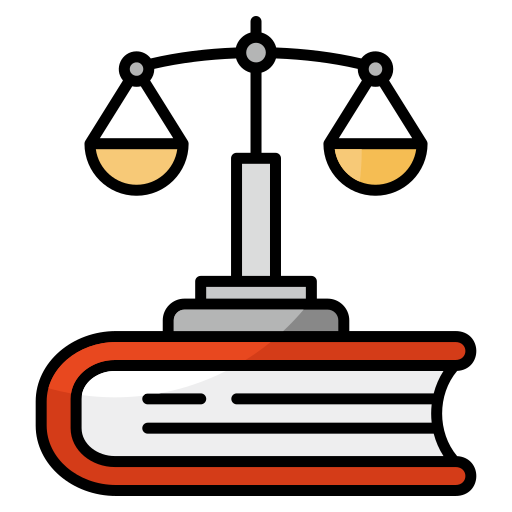}%
  \includegraphics[height=1pt]{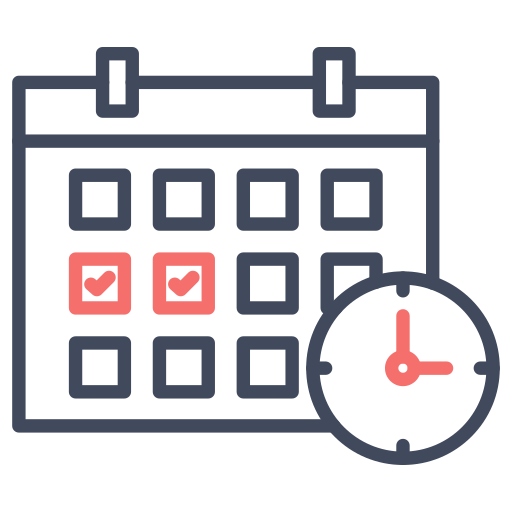}%
  \includegraphics[height=1pt]{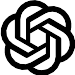}%
  \includegraphics[height=1pt]{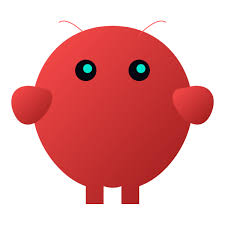}%
  \includegraphics[height=1pt]{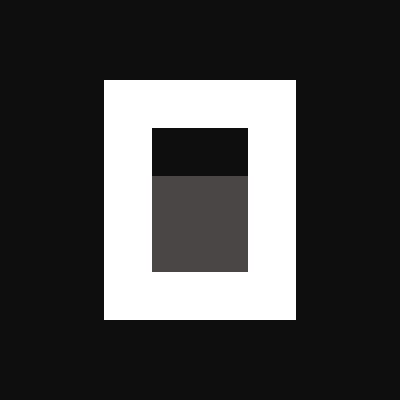}%
  \includegraphics[height=1pt]{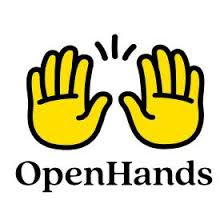}%
  \includegraphics[height=1pt]{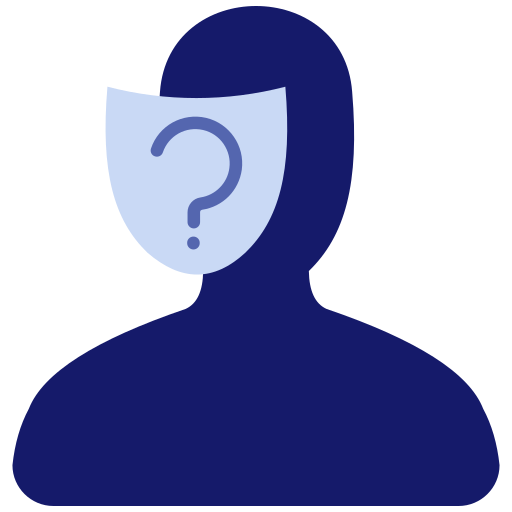}%
  \includegraphics[height=1pt]{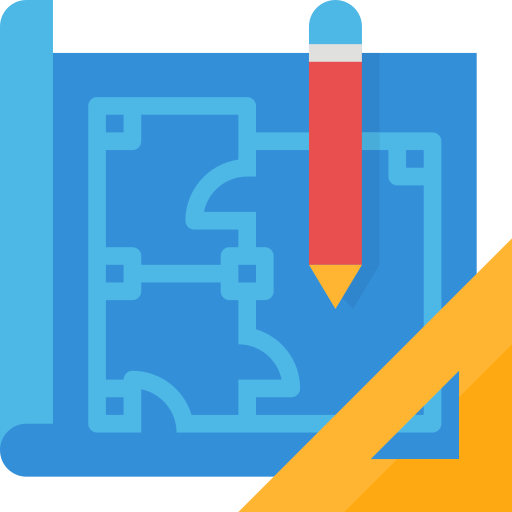}%
  \includegraphics[height=1pt]{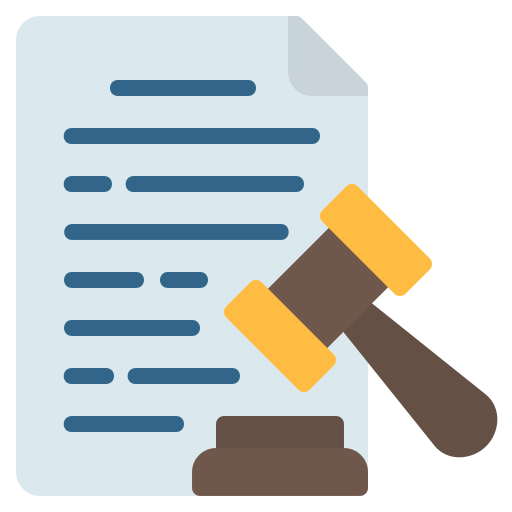}%
  \includegraphics[height=1pt]{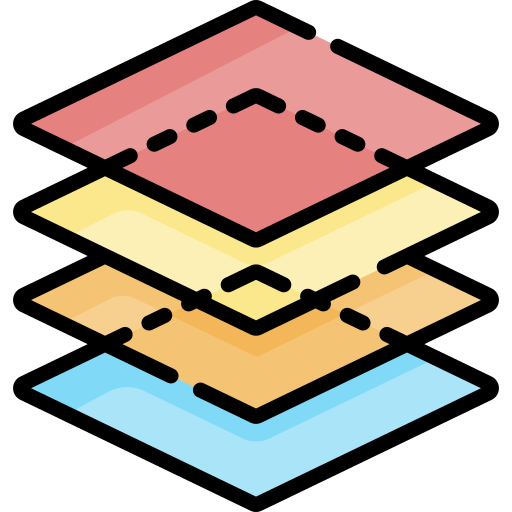}%
  \includegraphics[height=1pt]{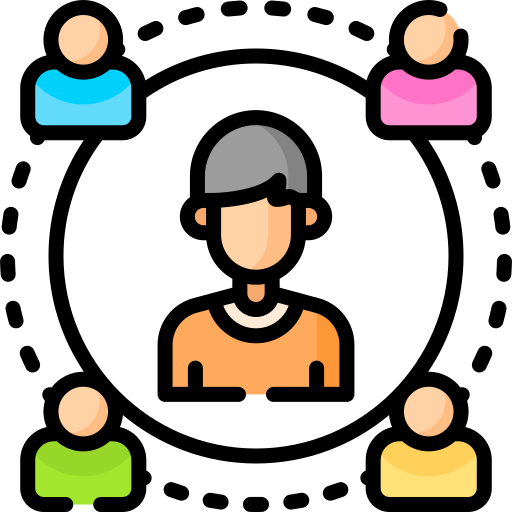}%
  \includegraphics[height=1pt]{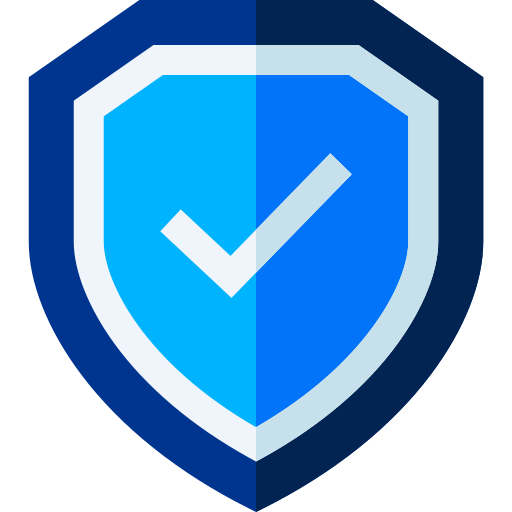}%
  \includegraphics[height=1pt]{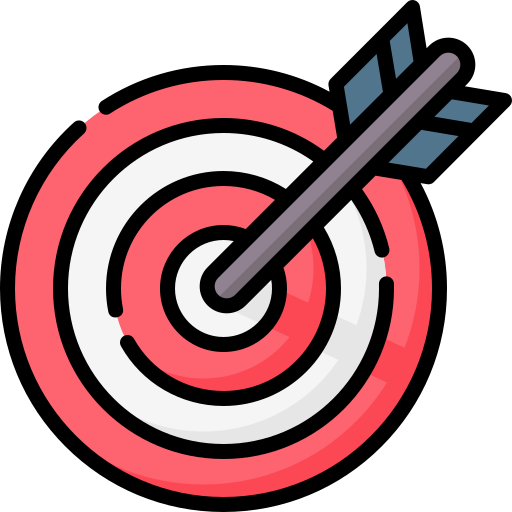}%
  \includegraphics[height=1pt]{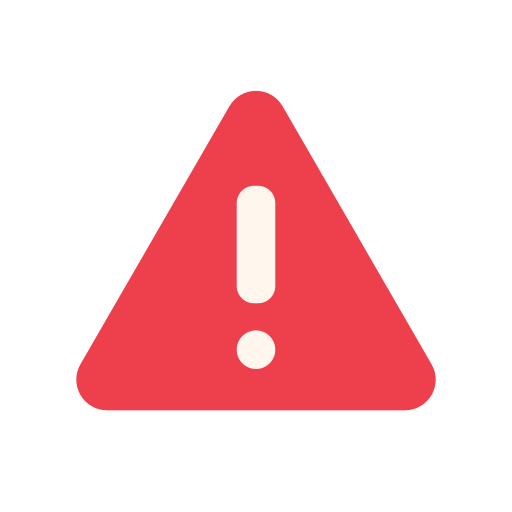}%
  \includegraphics[height=1pt]{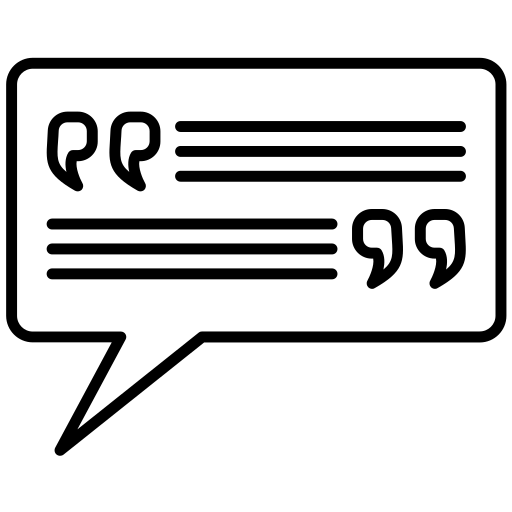}%
  \includegraphics[height=1pt]{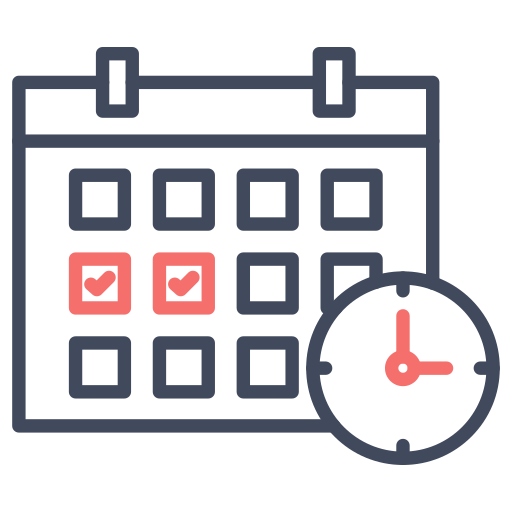}%
  \includegraphics[height=1pt]{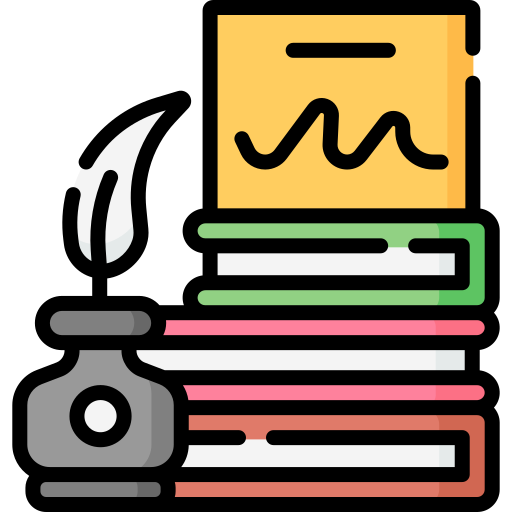}%
  \includegraphics[height=1pt]{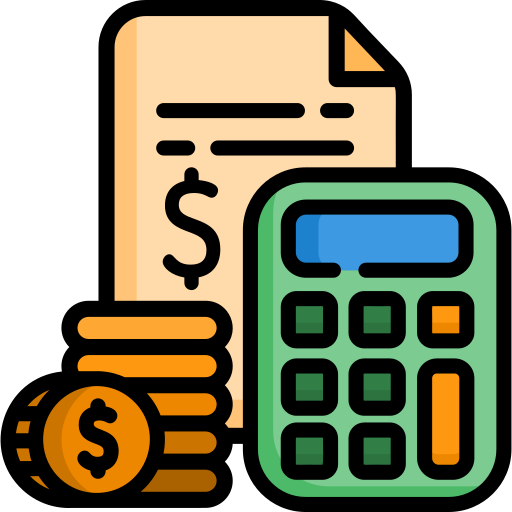}%
  \includegraphics[height=1pt]{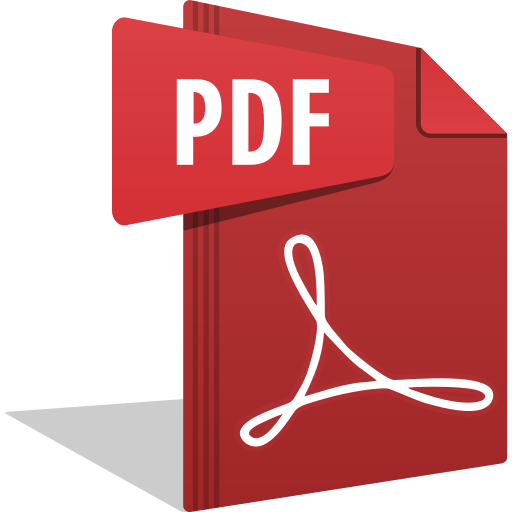}%
  \includegraphics[height=1pt]{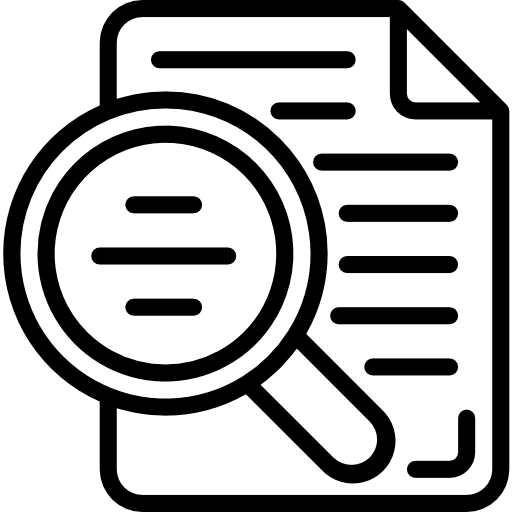}%
  \includegraphics[height=1pt]{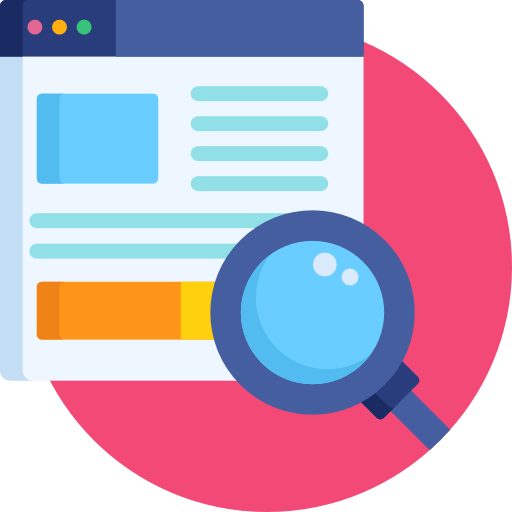}%
  \includegraphics[height=1pt]{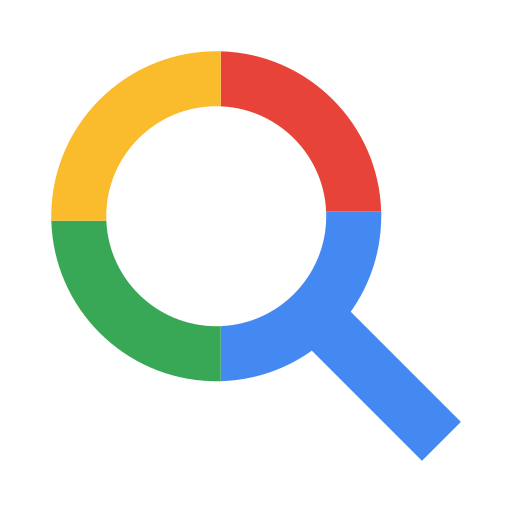}%
}%

\section{Related Work}
\label{sec:related}

\begin{figure}[!b]
\centering
\includegraphics[width=\textwidth]{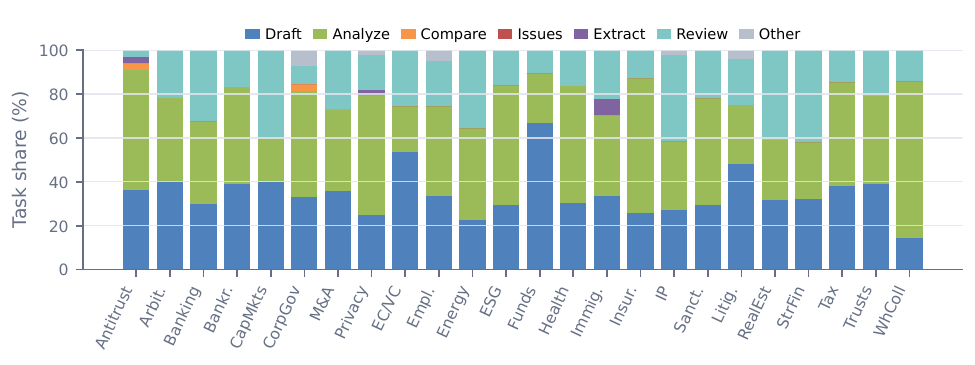}
\caption{\textbf{Primary task-archetype distribution by practice area
in the Harvey LAB dataset.} Bars are normalized within each of Harvey
LAB's $24$ practice areas; colors denote the primary task clusters.}
\label{fig:domain-worktype-distribution}
\end{figure}

\paragraph{Legal benchmarks and legal-agent evaluation.}
Legal NLP benchmarks mostly use bounded tasks: LegalBench
\citep{guha2023legalbench}, LexGLUE
\citep{chalkidis2022lexglue}, CUAD \citep{hendrycks2021cuad}, and
LawBench \citep{fei2023lawbench} probe reasoning, classification,
and clause extraction. Legal-agent systems
\citep{cui2023chatlaw,li2024legalagentbench,mantravadi2025legalwiz}
and source-grounding studies
\citep{pipitone2024legalbenchrag,dahl2024largelegalfictions,
kant2025robustlegalreasoning,hu2025legalfactuality} push toward
grounded deliverables. Closest to our setting, Harvey LAB
\citep{harvey2026lab} scores matter-style tasks with data rooms,
deliverables, all-pass rubrics, and a broad practice-area/archetype
mix (Figure~\ref{fig:domain-worktype-distribution}). Yet a large-scale
empirical analysis of these systems is still missing, which leaves the
true capabilities of state-of-the-art agents on open-ended legal matters
poorly understood.

\paragraph{Agent harnesses, externalization, and memory.}
The agent \emph{harness}---the runtime mediating workspace, tools,
traces, and output constraints---is the unit of long-horizon
evaluation. SWE-agent and SWE-bench
\citep{yang2024sweagent,jimenez2024swebench} make this explicit
for software engineering; GAIA, WebArena, and AgentBench
\citep{mialon2023gaia,zhou2023webarena,liu2023agentbench} provide
settings where traces and tool use can be studied. General-purpose
workspace agents (Codex-style, Claude Code, OpenCode) externalize
files, shell, search, and edits into an inspectable workspace.
Legal work additionally requires professional memory, provenance,
deadlines, citation form, and review signals as versionable artifacts.
Yet none of these harnesses is built for the legal vertical, leaving such
requirements as ad hoc prompt additions rather than first-class
architectural components.

\paragraph{Non-parametric agent learning.}
Non-parametric agent improvement edits prompts, programs, and
workflows rather than model weights. Prompt and
program optimization (APE, OPRO, PromptBreeder, APO, TextGrad,
PromptAgent, DSPy, GEPA;
\citealp{zhou2022ape,yang2023opro,fernando2023promptbreeder,
pryzant2023apo,yuksekgonul2024textgrad,wang2023promptagent,
khattab2023dspy,agrawal2025gepa}) and workflow search (DyLAN,
MetaGen, ADAS, AFlow, SEVerA;
\citealp{liu2023dylan,wang2026metagen,hu2024adas,zhang2024aflow,
banerjee2026severa}) treat the harness as a searchable optimization
target, while multi-agent systems
\citep{li2023camel,du2023debate,hong2023metagpt,qian2023chatdev,
wu2023autogen,chen2023agentverse} split roles across cooperating solvers.
Tool-use work
\citep{schick2023toolformer,qian2023creator,qin2023toolllm,
li2023apibank} and governance requirements
\citep{riedl2025aiagentslaw,mokander2023auditingllms,
wachter2024legaldutytruth} further motivate deterministic audit tools and
human-readable diffs. None of this work, however, targets the expensive,
leakage-sensitive setting of legal rollouts, nor combines legal domain
specialization with a self-evolving framework.

\section{The \textsc{Parthenon} Framework Architecture}
\label{sec:architecture}
\subsection{\textsc{Parthenon} architecture}
\label{subsec:parthenon-architecture}

\paragraph{Design principles.} \textsc{Parthenon} is built around three
principles. \textbf{Compatibility}: Codex, Claude Code, and related
workspace harnesses already supply the long-horizon execution needed
for legal work, including file management, tool calling, iterative
revision, and trace capture. The framework therefore wraps them with
legal state, deterministic tools, and procedural skills rather than
replacing them. \textbf{Legal specialization}: legal work carries hard
professional invariants that general harnesses do not enforce. A missed
deadline can constitute malpractice, an ungrounded figure can invalidate a
term, and an uncited authority is inadequate. \textsc{Parthenon} turns
these invariants into deterministic audit tools with mandatory
execution, covering source traceability, date and number reconciliation,
deliverable compliance, and issue-lifecycle closure. This replaces
implicit model memory with an enforceable harness contract.
\textbf{Dynamism}: agent failures recur in the same procedural
categories across models and practice areas, as documented in
\S\ref{sec:experiments}. A self-evolving loop therefore turns scored
failures into reviewable harness diffs, mirroring how firms update
checklists instead of retraining personnel. This follows the lineage of
tool-using and self-improving agents \citep{schick2023toolformer,
qin2023toolllm,shinn2023reflexion,madaan2023selfrefine,zhao2023expel}
but adds structural anti-leakage.

\subsubsection{Model layer}
The Model layer is a \textbf{pluggable capability provider}, not a
fixed binding. Initial LAB results show no single frontier model
dominates: different families lead in different legal sub-domains
\citep{harvey2026labresults}, and binding the framework to one
model would convert those blind spots -- weak temporal arithmetic,
citation hallucination, thin practice-area coverage -- into
system-level failures. A thin capability interface lets a
deployment route matters by practice area, task shape, quality,
or cost while the rest of the stack stays stable; because the model
and agent harness stay fixed, \textsc{Parthenon}'s lift can be measured
cleanly (\S\ref{subsec:patterns}).

\subsubsection{Harness layer}
The Harness supplies the \textbf{observable execution contract}:
matter workspace, source/tool access, mediated read/search/execute/
write/validate/edit actions, and trace capture. LAB trace analysis
shows that in-task behavior -- searching before drafting,
validating after -- correlates with outcomes
\citep{harvey2026labresults}, so the harness is where model
capability becomes legal work, not merely a wrapper.
\textsc{Parthenon} treats it as pluggable: Codex, Claude Code,
OpenCode, OpenHands, or basic legal-native agents can occupy it
so long as they expose workspace, tools, deliverable channel, and
traces. The legal contribution wraps that runtime, giving
\textbf{harness-agnostic specialization}: the legal layer runs
across runtimes, and runtime-specific behavior stays measurable.
Without this separation, an improvement could be a workspace-ergonomic
or a legal-control change, and the harness could not migrate when the
vendor changes.

\subsubsection{Agent layer}
The Agent layer defines \textbf{role and information boundaries}
on top of the harness. Reliability comes from separating three
functions that must not share information -- drafting the work
product, evaluating it, and proposing harness edits -- not from
adding agents to drafting; the matter-facing worker stays a single
solver. Collapsing these lets prompt-optimization loops leak: a system that both drafts and
grades against the rubric can eventually memorize it. The split is
therefore central to attribution and anti-leakage. The
\textbf{solver} drafts and self-audits given task, sources,
skill, and tools; the \textbf{evaluator} scores the finished work
against the rubric outside the solver's context, so its judgments
cannot leak into draft text; the \textbf{learner} sees only
redacted traces and aggregate signals and proposes task-agnostic
edits to Knowledge, Tools, or Skills, subject to automated rejection gates.
\S\ref{sec:learning} specifies the loop over batches.

\begin{figure}[t]
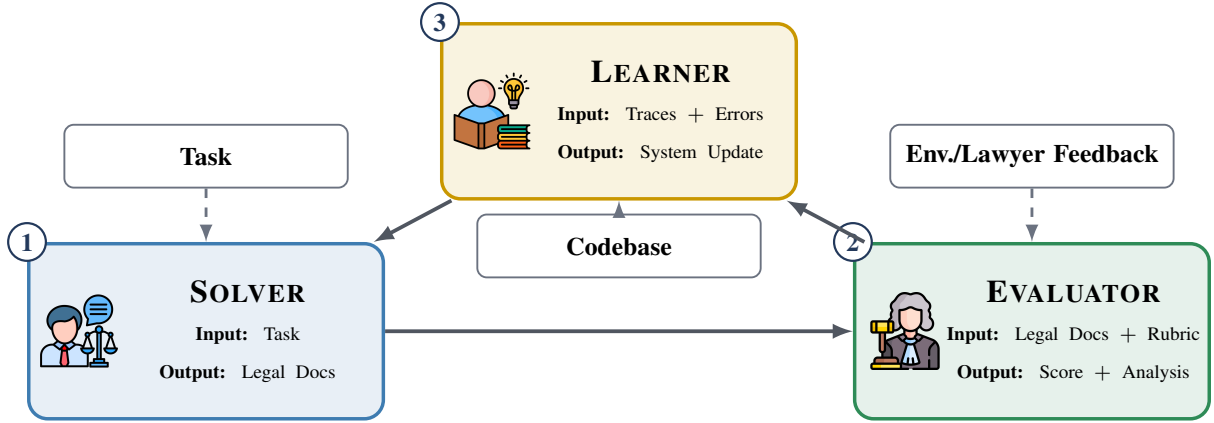

\centering
\resizebox{\textwidth}{!}{%
\begin{tikzpicture}[
  x=1cm, y=1cm,
  agent/.style={
    rounded corners=6pt,
    draw=#1,
    fill=#1!12,
    line width=1.1pt,
    minimum width=3.95cm,
    minimum height=1.95cm,
    text width=3.65cm,
    align=center,
    font=\small\bfseries
  },
  extbox/.style={
    rounded corners=3pt,
    draw=plline,
    fill=white,
    line width=0.6pt,
    minimum width=3.15cm,
    minimum height=0.7cm,
    text width=2.9cm,
    align=center,
    font=\scriptsize
  },
  loop/.style={
    ->, >=latex,
    line width=1.2pt,
    draw=plline!75!black
  },
  ext/.style={
    ->, >=latex,
    line width=0.9pt,
    draw=plline,
    dashed
  },
  edgelbl/.style={
    fill=white,
    inner sep=2pt,
    font=\scriptsize\itshape,
    text=plnavy
  },
  numbadge/.style={
    circle,
    draw=plnavy,
    fill=white,
    line width=0.7pt,
    minimum size=0.42cm,
    inner sep=0pt,
    font=\scriptsize\bfseries,
    text=plnavy
  }
]
\definecolor{plnavy}{HTML}{23395B}
\definecolor{plblue}{HTML}{3E7CB1}
\definecolor{plgreen}{HTML}{2E8B57}
\definecolor{plgold}{HTML}{C99700}
\definecolor{plred}{HTML}{B35C44}
\definecolor{plline}{HTML}{6B7280}

% Three agents arranged as a triangle (compressed to ~3:1 aspect)
\node[agent=plblue] (solver) at (2.6,1.10) {};
\node[anchor=west, inner sep=0pt] at ([xshift=4pt]solver.west)
  {\includegraphics[height=0.86cm]{figures/logos/agent/solver.png}};
\node[anchor=west, text width=2.9cm, align=center, inner sep=0pt,
      font=\small\bfseries] at ([xshift=1.0cm]solver.west)
  {\textsc{Solver}\\[5pt]\tiny\mdseries
   \textbf{Input:} Task\\\tiny\mdseries
   \textbf{Output:} Legal Docs};
\node[numbadge] at (solver.north west) {1};

\node[agent=plgold] (learner) at (7.2,3.55) {};
\node[anchor=west, inner sep=0pt] at ([xshift=4pt]learner.west)
  {\includegraphics[height=0.86cm]{figures/logos/agent/learn.png}};
\node[anchor=west, text width=2.9cm, align=center, inner sep=0pt,
      font=\small\bfseries] at ([xshift=1.0cm]learner.west)
  {\textsc{Learner}\\[5pt]\tiny\mdseries
   \textbf{Input:} Traces $+$ Errors\\\tiny\mdseries
   \textbf{Output:} System Update};
\node[numbadge] at (learner.north west) {3};

\node[agent=plgreen] (evaluator) at (11.8,1.10) {};
\node[anchor=west, inner sep=0pt] at ([xshift=4pt]evaluator.west)
  {\includegraphics[height=0.86cm]{figures/logos/agent/judge.png}};
\node[anchor=west, text width=2.9cm, align=center, inner sep=0pt,
      font=\small\bfseries] at ([xshift=1.0cm]evaluator.west)
  {\textsc{Evaluator}\\[5pt]\tiny\mdseries
   \textbf{Input:} Legal Docs $+$ Rubric\\\tiny\mdseries
   \textbf{Output:} Score $+$ Analysis};
\node[numbadge] at (evaluator.north west) {2};

% External inputs placed around the triangle
\node[extbox] (task)     at (2.6,3.05)   {\textbf{Task}};
\node[extbox] (feedback) at (11.8,3.05)  {\textbf{Env./Lawyer Feedback}};
\node[extbox] (codebase) at (7.2,2.05)   {\textbf{Codebase}};

\draw[ext] (task)     -- (solver);
\draw[ext] (feedback) -- (evaluator);
\draw[ext] (codebase) -- (learner);

% Directed learning loop
\draw[loop] (solver) -- (evaluator);
\draw[loop] (evaluator) -- (learner);
\draw[loop] (learner) -- (solver);
\end{tikzpicture}%
}

\caption{\textbf{Self-evolving loop in \textsc{Parthenon}.}\ The
\textsc{solver} drafts legal documents, the \textsc{evaluator} scores
them against the rubric, and the \textsc{learner} turns redacted
traces and errors into harness updates over the editable codebase
(tools, knowledge, skills). Redaction between evaluator and learner
(\S\ref{sec:learning}) enforces single-pass anti-leakage.}
\label{fig:solver-evaluator-learner}
\end{figure}

\subsubsection{Knowledge layer}
The Knowledge layer stores \textbf{durable legal memory as data,
not prompt text}, so a change to a deadline rule, authority, schema,
or synonym is a reviewable harness edit rather than a hidden model
behavior. Knowledge artifacts do not execute checks; they give Tools
structured content to retrieve and validate, and Skills stable
legal objects to reference -- institutional memory rather than
ad hoc prompt content. Table~\ref{tab:knowledge-families} lists the six families
used in our deployment. The hard boundary is that Knowledge is
\textbf{not an answer bank}: a permissible edit adds a general
authority, deadline rule, calendar, synonym, or work-product schema,
but never a task id, source fact, client identity, rubric phrase, or
answer from the batch that exposed the failure. Knowledge must be
reusable across matters, otherwise it is benchmark leakage or a
privacy risk.

\begin{table}[H]
\centering
\small
\setlength{\tabcolsep}{7pt}
\begin{tabularx}{\textwidth}{@{}p{2.8cm}X@{}}
\toprule
\textbf{Family} & \textbf{Contents} \\
\midrule
Statute catalog
  & Statutory citations, regulations, procedural rules, case law, ethics rules, and secondary sources with keyword and jurisdiction metadata. Covers 13 jurisdictions (US federal/state, EU, UK, international). \\[4pt]
Window catalog
  & Deadline and notice windows keyed by regime and trigger event, with calendar-day vs.\ business-day classification and rolling rules. Spans 66 deadline regimes across 15 legal domains. \\[4pt]
Deliverable catalog
  & Required-section skeletons for work-product validation, organized by document type and practice area. Covers 25 practice areas from M\&A to immigration. \\[4pt]
Holiday calendars
  & Jurisdiction-specific holiday sets for accurate business-day arithmetic covering federal courts, NYSE, IRS, bankruptcy courts, and international venues. Includes US, Europe, Asia, and Americas variants. \\[4pt]
Legal synonyms
  & Canonical concept keys mapped to alias terms for robust cross-document term matching. Applied cross-domain to normalize terminology across practice areas. \\[4pt]
Inference rules
  & Regex dispatch cues that map filename and text signals to document types and retrieval triggers. Organized across 25 practice areas with universal fallback patterns. \\
\bottomrule
\end{tabularx}
\caption{\textbf{Knowledge families in \textsc{Parthenon}.}
The Knowledge layer is organized into six reusable families totaling
over 2{,}300 entries. All entries are general-purpose legal objects
(authorities, deadlines, schemas, calendars, synonyms, dispatch rules)
and must not encode matter-specific facts, client identities, or
benchmark-derived answers.}
\label{tab:knowledge-families}
\end{table}

\subsubsection{Tools layer}
The Tools layer turns recurring legal requirements into
\textbf{deterministic, interpretable operations}. Whenever a requirement
reduces to inspection, retrieval, parsing, arithmetic, date logic, or
citation and deliverable checks, \textsc{Parthenon} makes it an executable
tool rather than unstated prompt intent. This matters because the most
common failures (\S\ref{subsec:trajectory}) are mechanical but
material -- a wrong deadline, omitted source, unsupported citation, lost
amount, malformed redline -- each making an otherwise plausible answer
unreleasable. The surface follows the matter lifecycle
(Table~\ref{tab:tools-catalogue-detail}): pre-draft tools canonicalize
sources into typed matter state; retrieval tools query the Knowledge
layer; inspectors parse spreadsheets, documents, emails, and summaries;
computation tools reconcile numbers and dates; post-draft release gates
audit citation grounding, coverage, numeric/temporal consistency, and
deliverable shape. Skills decide when a tool is mandatory; Tools decide
what can be checked without trusting model memory.

\begin{table}[H]
\centering
\footnotesize
\setlength{\tabcolsep}{4pt}
\renewcommand{\arraystretch}{1.1}
\begin{tabularx}{\textwidth}{@{}L{3.2cm}Y@{}}
\toprule
\textbf{Stage} & \textbf{Contents} \\
\midrule
Source inspection & Spreadsheet, Word, PPT, email inspectors build a typed pre-draft inventory. \\
Knowledge retrieval & Authority, legal-window, and deliverable-schema search supply explicit inputs. \\
Audit \& computation & Citation grounding, number/date reconciliation, deadline arithmetic post-draft. \\
Release gates & Skeleton builder + audit dispatcher produce one reviewable release report. \\
\bottomrule
\end{tabularx}
\caption{Tools-layer: $14$ deterministic, agent-callable capabilities,
summarized here into the four legal-work stages they serve and
enumerated individually in Appendix~\ref{app:tools-detail}. Each
stage enforces operational contracts that do not rely on model
memory; the release gate consolidates citation, coverage, and
deliverable checks into one reviewable report before any work product
leaves the matter.}
\label{tab:tools-catalogue-detail}
\end{table}

\subsubsection{Skills layer}
The Skills layer is the most task-facing content layer: it turns
recurring legal work into \textbf{rubric-blind procedural
plans}. A skill is selected for a matter class, not an answer, and
specifies triage, required source coverage and issue lifecycle,
mandatory tools, deliverable form, and the self-audit that blocks
finalization. This layer is where empirical failures become reusable
procedure: deliverable substitution $\to$ work-product identity
checking; numeric/date loss $\to$ mandatory reconciliation;
single-pass drafting $\to$ explicit decomposition; missing
recommendations $\to$ issue-closure checklists. Each skill follows a
seven-part scaffold (Table~\ref{tab:skills-catalogue}; an example
appears in Appendix~\ref{tab:skill-anatomy}) -- triage,
failure modes corrected, legal frameworks, analytical scaffolds,
relationships, output structure, anti-leakage checklist -- and the
final checklist forbids task ids, rubric phrases, client identities,
deal amounts, dates, or private quotations from the batch that
exposed the failure. Skills are versioned artifacts the learner can
edit, promoted only when the edit passes the acceptance gates.

\begin{table}[H]
\centering
\footnotesize
\setlength{\tabcolsep}{4pt}
\renewcommand{\arraystretch}{1.1}
\begin{tabularx}{\textwidth}{@{}L{4.2cm}Y@{}}
\toprule
\textbf{Skill family} & \textbf{Procedure $\to$ required controls} \\
\midrule
Drafting \& production & Instantiate schema and close clauses $\to$ skeletons + release gates. \\
Issue identification & Inventory facts $\to$ rule $\to$ severity $\to$ action $\to$ coverage + lifecycle gate. \\
Comparison \& synthesis & Define axes, reconcile across documents $\to$ cross-doc map. \\
Extraction \& reconstruction & Enumerate fields, preserve row granularity $\to$ inspectors + missing-field audit. \\
Scenario application & Decompose, bind facts to rule elements $\to$ retrieval + fact-to-element matrix. \\
Analysis \& research & Standard + authorities $\to$ citation grounding + source audit. \\
Other specialized & Targeted playbooks $\to$ task-specific coverage gate. \\
\bottomrule
\end{tabularx}
\caption{Skills-layer catalogue: $1{,}251$ task-routed procedural
skills. A skill is selected by matter class, never by answer, and
encodes a rubric-blind plan -- triage, coverage, mandatory tools,
deliverable form, self-audit gate. Edits are versioned diffs, turning
failure modes into reusable procedure rather than rubric memorization.}
\label{tab:skills-catalogue}
\end{table}

\subsection{Self-Evolving Loop: Gated Harness Optimization}
\label{sec:learning}

\textsc{Parthenon} does not change the model; it changes the harness
that each subsequent model call inhabits. At step $t$, the foundation
model $M$ and workspace harness $H$ are fixed, while the editable
harness
\[
S_t=(K_t,T_t,G_t,A_t)
\]
holds durable Knowledge $K_t$, the deterministic Tools surface $T_t$,
the Skills library $G_t$, and role prompts + audit gates $A_t$. A
learning step is therefore a versioned, human-readable harness
commit, not a parameter update.

\subsubsection{Interfaces and update rule}

A matter is $m=(q,X,d,C)$ with task brief $q$, source set $X$,
deliverable spec $d$, and hidden rubric $C$ (never given to the
solver). The three agent roles produce three signals:
\begin{equation}
\text{\textbf{Solver:}}\quad
(\hat y,\tau)=\mathrm{Solve}_M(q,X\mid H,S_t,d),
\label{eq:solver}
\end{equation}
\begin{equation}
\text{\textbf{Evaluator:}}\quad
z=\mathrm{Evaluate}_J(\hat y\mid q,d,C),
\label{eq:evaluator}
\end{equation}
\begin{equation}
\text{\textbf{Learner:}}\quad
\widehat{\Delta}_t=\mathrm{Learn}_L(E_t,Z_t,\mathrm{Repo}(S_t)),
\label{eq:learner}
\end{equation}
where $E_t$ holds redacted failure trajectories and $Z_t$ the judge
feedback over a without-replacement minibatch $B_t$; the trace $\tau$
serves as supervision signal and is never transcribed directly into the harness. The harness then advances
as
\begin{equation}
S_{t+1}=S_t\oplus\widehat{\Delta}_t.
\label{eq:update}
\end{equation}
The analogy to SGD is operational: the harness is discrete and
reviewable, so $\oplus$ denotes the accepted effect of a bounded diff
after rejection gates, not blind application of learner text.

\paragraph{Anti-leakage and rejection gates.} Leakage protection is
structural. The solver receives task, sources, deliverable spec, and
harness but never criterion ids, titles, or match criteria; the
evaluator is post-hoc and never feeds back during drafting; the
learner sees only an abstract task shape, file names, bucketed
tool-use counts, a truncated final solver message, aggregate
pass/fail counts, and de-identified failure reasons. The learner must emit
matter-independent edits. A bounded learner diff is admitted only if
(i) the feedback is general enough to be edited at all, (ii) any
tool code compiles and passes static safety checks, and (iii) the
candidate harness strictly improves the accepted per-task pass
rate. Otherwise the previous harness is reinstated and the rejected
candidate is logged.

\subsubsection{What the loop edits}

Each $\widehat{\Delta}_t$ is attributable to one harness surface:
\textbf{Tools} (Table~\ref{tab:tools-catalogue-detail}) turn
mechanical legal invariants -- dates, deadlines, amounts, coverage,
citation presence, deliverable shape -- into audited code rather than
unstated prompt intent; \textbf{Skills}
(Table~\ref{tab:skills-catalogue}) are rubric-blind work plans naming
inventory, issue lifecycle, mandatory tools, deliverable identity,
and the finalization audit but never an answer; \textbf{Knowledge}
stores durable legal objects (authorities, windows, schemas,
calendars, synonyms) as data retrieved through tools. The hard
boundary is \emph{procedure versus conclusion}: a skill that encodes
a specific conclusion from one matter is a memorized answer and a
client-data leakage vector, so a rubric-blind audit -- given only
task title and skill text -- rewrites task-specific content into
generic procedure before commit. A vocabulary check over the
$1{,}251$ skills confirms that only domain-general legal terms are shared
with rubric criteria. Empirical lift, concrete learner edits, and the
hard-$10$ optimization ledger appear in \S\ref{subsec:ablations}.

\section{Experiments and Analysis}
\label{sec:experiments}

\subsection{Experimental setup and results}
\label{subsec:baselines}

\paragraph{Benchmark and execution modes.} Harvey LAB contains
$1{,}251$ matters across $24$ practice areas (median $7$ source
documents and $57$ criteria per matter;
Figure~\ref{fig:domain-worktype-distribution}). Table~\ref{tab:baseline-summary}
reports four execution families: direct prompting (API), a basic
legal-native harness, and the Codex and Claude Code workspace harnesses.
Each baseline is paired with a same-solver \textsc{Parthenon} column, so
the lift isolates the legal harness rather than a model upgrade. We report
two metrics: \emph{criterion accuracy}, the share of all rubric
criteria passed, and \emph{all-pass}, the stricter share of matters that
pass \emph{every} criterion.

\paragraph{Protocol.} All full-corpus cells share the task universe
($1{,}251$ tasks), judge, and deliverable-shape
conversion; solver settings are fixed within each paired comparison. No
solver reads judge outputs or failed-criterion text during execution
(\S\ref{sec:learning}'s anti-leakage rules). Each paired comparison holds
the solver and its workspace fixed and adds only the optimized
\textsc{Parthenon} harness. All reasoning models run at the
\texttt{medium} reasoning-effort setting.

\begin{table}[H]
\centering
\scriptsize
\setlength{\tabcolsep}{2.2pt}
\setlength{\abovecaptionskip}{3pt}
\setlength{\belowcaptionskip}{0pt}
\renewcommand{\arraystretch}{1.00}
\resizebox{\textwidth}{!}{%
\begin{tabular}{@{}l C{1.05cm}|C{1.05cm}||C{1.05cm}|C{1.05cm}|C{1.05cm}|C{1.05cm}||C{1.05cm}|C{1.05cm}|C{1.05cm}|C{1.05cm}@{}}
\toprule
Model & \multicolumn{2}{c||}{GPT-5.4-mini} & \multicolumn{2}{c|}{GPT-5.4-mini} & \multicolumn{2}{c||}{GPT-5.5} & \multicolumn{2}{c|}{Haiku 4.5} & \multicolumn{2}{c}{Sonnet 4.6} \\
Harness & API & Basic & Codex & \textsc{Parth.} & Codex & \textsc{Parth.} & Claude & \textsc{Parth.} & Claude & \textsc{Parth.} \\
\midrule
All-pass & 0.40\%& 0.16\%& 1.12\%& \textbf{3.36\%}& 3.76\%& \textbf{10.95\%}& 1.12\%& \textbf{3.04\%}& 11.83\%& \textbf{11.99\%} \\
Criteria acc. & 56.5\%& 60.7\%& 68.2\%& \textbf{82.0\%}& 79.8\%& \textbf{89.9\%}& 71.6\%& \textbf{77.8\%}& 82.8\%& \textbf{90.2\%} \\
\midrule
EC/VC& 47.4& 61.0& 56.7& \textbf{82.3}& 69.8& \textbf{89.6}& 66.3& \textbf{79.9}& 76.5& \textbf{89.9} \\
Funds& 44.1& 67.9& 65.1& \textbf{84.8}& 72.3& \textbf{90.9}& 67.6& \textbf{80.0}& 74.5& \textbf{90.1} \\
Trusts& 55.1& 62.8& 65.2& \textbf{82.0}& 74.6& \textbf{89.7}& 71.7& \textbf{81.3}& 76.5& \textbf{92.4} \\
Real Estate& 63.3& 62.8& 66.2& \textbf{82.9}& 75.4& \textbf{91.1}& 70.5& \textbf{79.7}& 80.2& \textbf{90.3} \\
Structured Fin& 54.2& 62.9& 68.9& \textbf{81.1}& 76.0& \textbf{91.2}& 67.7& \textbf{75.1}& 86.4& \textbf{90.6} \\
Immigration& 57.2& 54.2& 63.1& \textbf{77.7}& 77.8& \textbf{86.1}& 59.5& \textbf{79.6}& 75.2& \textbf{86.5} \\
Capital Mkts& 44.4& 64.2& 68.8& \textbf{80.8}& 78.3& \textbf{89.9}& 64.8& \textbf{74.2}& 82.0& \textbf{88.6} \\
Litigation& 58.0& 59.8& 64.4& \textbf{77.7}& 78.7& \textbf{87.7}& 68.6& \textbf{75.4}& 82.4& \textbf{88.4} \\
Tax& 50.4& 48.0& 68.7& \textbf{79.8}& 78.9& \textbf{87.0}& 66.4& \textbf{69.8}& 83.9& \textbf{84.1} \\
Antitrust& 59.9& 54.7& 67.4& \textbf{81.7}& 79.5& \textbf{87.8}& 72.7& \textbf{80.6}& 79.0& \textbf{85.9} \\
Employment& 61.2& 60.8& 66.2& \textbf{82.9}& 79.5& \textbf{92.2}& 70.9& \textbf{81.9}& 80.9& \textbf{91.4} \\
IP& 61.4& 64.3& 68.9& \textbf{84.1}& 80.6& \textbf{91.8}& 73.9& \textbf{79.2}& 85.7& \textbf{91.5} \\
ESG& 62.4& 56.2& 69.0& \textbf{78.0}& 80.7& \textbf{88.1}& 72.3& \textbf{75.7}& 85.2& \textbf{89.9} \\
Banking& 53.1& 59.4& 70.2& \textbf{79.7}& 80.8& \textbf{89.6}& 70.4& \textbf{70.8}& 81.6& \textbf{89.4} \\
Healthcare& 58.0& 59.6& 64.0& \textbf{84.3}& 81.4& \textbf{89.6}& 69.3& \textbf{76.6}& 82.1& \textbf{91.2} \\
Bankruptcy& 58.3& 53.6& 71.4& \textbf{81.9}& 81.5& \textbf{89.1}& 71.2& \textbf{78.9}& 81.9& \textbf{90.6} \\
Privacy& 66.9& 65.4& 69.3& \textbf{84.8}& 81.6& \textbf{92.2}& 77.4& \textbf{85.5}& 87.2& \textbf{92.9} \\
Corp Gov& 60.6& 54.8& 70.7& \textbf{81.1}& 82.0& \textbf{89.5}& 74.6& \textbf{79.7}& 86.8& \textbf{90.7} \\
Sanctions& 61.0& 52.2& 68.3& \textbf{81.8}& 82.8& \textbf{89.2}& 72.1& \textbf{76.3}& 87.0& \textbf{89.5} \\
Arbitration& 70.0& 63.1& 72.6& \textbf{85.4}& 83.2& \textbf{90.3}& \textbf{83.9}& 80.9& 86.4& \textbf{94.8} \\
White Collar& 51.4& 54.1& 73.5& \textbf{78.8}& 83.3& \textbf{86.3}& 63.9& \textbf{68.1}& \textbf{87.3}& 86.5 \\
Corp M\&A& 52.1& 63.2& 72.8& \textbf{80.5}& 83.7& \textbf{89.3}& 73.5& \textbf{75.0}& 83.8& \textbf{89.1} \\
Insurance& 60.5& 56.3& 73.3& \textbf{83.0}& 84.6& \textbf{91.3}& 74.8& \textbf{75.3}& 87.0& \textbf{91.5} \\
Energy& 58.8& 63.3& 64.2& \textbf{85.6}& 88.6& \textbf{92.8}& 75.3& \textbf{76.9}& 90.8& \textbf{93.8} \\
\bottomrule
\end{tabular}
}
\caption{Baselines and \textsc{Parthenon} on Harvey LAB. All-pass is
strict matter pass rate; other rows are criterion accuracy. Bold marks
the better same-solver paired cell; API/Basic are unpaired. Domains
are sorted by Codex/GPT-5.5 accuracy.}
\label{tab:baseline-summary}
\end{table}

\paragraph{Baseline pattern and framework effect.} The first regularity
is that a stronger base model raises \emph{criterion} accuracy but not
\emph{matter}-level completion. Codex climbs from $68.2$ to $79.8\%$ as the
solver moves from GPT-5.4-mini to GPT-5.5, and Claude Code with Sonnet
4.6 reaches $82.8\%$. Yet even the strongest baseline passes every
criterion on barely one matter in eight ($11.8\%$). All-pass is a
conjunction over dozens of fact, authority, deadline, and deliverable
conditions, so a few points of average accuracy cannot satisfy them
jointly. The bottleneck is therefore systemic rather than a matter of
model scale.

The second regularity is that, with the model and agent harness fixed,
\textsc{Parthenon} alone delivers a lift comparable to a model upgrade at
every tier:
$+13.8/+10.2/+7.4$\,pp, reaching $82.0/89.9/90.2\%$ on mini, GPT-5.5, and
Sonnet. The gain is thus a property of the harness rather than of any
base model. The lift is largest where the base solver is weakest, which
narrows the gap between solvers. Strict completion roughly triples on the
weaker solvers but barely moves on Sonnet.
\S\ref{subsec:trajectory} breaks that remaining error down by type.
                    % 4.1 setup, protocol, results
\subsection{Baseline error analysis}
\label{subsec:trajectory}

\paragraph{Agents are active but incomplete.} Outcome scores indicate
whether a harness fails, not why. We audit the full Codex/GPT-5.5 result
cell: $1{,}251$ scored runs with paired native transcripts, final
artifacts, and process metrics. The cell passes about $79.8\%$ of all
rubric criteria pooled, and the failed criteria anchor the
Codex baseline column of
Table~\ref{tab:residual-error-breakdown}. We map judge rationales to
ten scripted error classes and parse traces for shell activity, agent
messages, and length. The traces are dense: a typical run issues many
shell commands and produces a long trajectory before emitting a finished
artifact. What the run lacks is closure. Formulas, source-specific facts,
required sections, and rule conditions that were available in the data
room do not reach the final work product, because the harness carries no
release signal for ``all required sources, numbers, authorities, and
deliverables satisfied.'' Effort is not the bottleneck; verification is.

\paragraph{Five classes carry most of the error.} Five error classes
dominate: missing facts, numbers/dates, legal-rule use, deliverable form,
and coverage. Together they account for nearly two thirds of the failed
criteria, with the remainder split across long-tail or unnamed causes
(Table~\ref{tab:residual-error-breakdown}, which gives the per-cause rate
for every cell). These are mechanical but material misses, and they are
diagnostic. Workspace agents inspect many files yet keep no typed matter
state, coverage manifest, or release gate, which is exactly what these
classes demand. \textsc{Parthenon} adds these controls and lowers the
total error at every solver, with the form- and grounding-class errors
falling fastest.

\begin{table}[t]
\centering
\scriptsize
\setlength{\tabcolsep}{2.2pt}
\renewcommand{\arraystretch}{1.05}
\resizebox{\textwidth}{!}{%
\begin{tabular}{@{}l C{1.05cm}|C{1.05cm}||C{1.05cm}|C{1.05cm}|C{1.05cm}|C{1.05cm}||C{1.05cm}|C{1.05cm}|C{1.05cm}|C{1.05cm}@{}}
\toprule
Model & \multicolumn{2}{c||}{GPT-5.4-mini} & \multicolumn{2}{c|}{GPT-5.4-mini} & \multicolumn{2}{c||}{GPT-5.5} & \multicolumn{2}{c|}{Haiku 4.5} & \multicolumn{2}{c}{Sonnet 4.6} \\
Harness & API & Basic & Codex & \textsc{Parth.} & Codex & \textsc{Parth.} & Claude & \textsc{Parth.} & Claude & \textsc{Parth.} \\
\midrule
Missing facts/entities& 11.4 & 10.3 & 7.2 & \textbf{4.7} & 4.7 & \textbf{2.7} & 7.3 & \textbf{5.4} & 3.8 & \textbf{2.5} \\
Numbers/dates& 10.5 & 10.1 & 5.5 & \textbf{3.2} & 3.3 & \textbf{1.7} & 6.4 & \textbf{3.8} & 2.8 & \textbf{1.7} \\
Legal-rule use& 4.0 & 3.5 & 2.7 & \textbf{1.5} & 2.0 & \textbf{0.8} & 2.7 & \textbf{1.8} & 1.5 & \textbf{0.8} \\
Deliverable/format& 3.5 & 2.9 & 2.0 & \textbf{1.0} & 1.7 & \textbf{0.6} & 2.3 & \textbf{1.2} & 1.6 & \textbf{0.6} \\
Coverage& 2.2 & 1.9 & 2.3 & \textbf{1.4} & 1.2 & \textbf{0.7} & \textbf{1.4} & 1.6 & 1.1 & \textbf{0.8} \\
Materiality/impact& 1.8 & 1.8 & 1.3 & \textbf{0.6} & 0.8 & \textbf{0.3} & 1.2 & \textbf{0.8} & 0.5 & \textbf{0.3} \\
Source grounding& 1.6 & 1.6 & 1.1 & \textbf{0.5} & 0.8 & \textbf{0.3} & 1.1 & \textbf{0.7} & 0.5 & \textbf{0.3} \\
Remedial action& 2.2 & 1.9 & 1.0 & \textbf{0.6} & 0.7 & \textbf{0.3} & 1.6 & \textbf{0.7} & 0.5 & \textbf{0.3} \\
Cross-doc synthesis& 0.1 & 0.1 & 0.1 & \textbf{0.0} & 0.0 & \textbf{0.0} & 0.1 & \textbf{0.0} & 0.0 & \textbf{0.0} \\
Other& 6.2 & 5.3 & 8.6 & \textbf{4.4} & 5.0 & \textbf{2.6} & \textbf{4.3} & 6.1 & 4.9 & \textbf{2.6} \\
\midrule
Total residual& 43.5 & 39.3 & 31.8 & \textbf{18.0} & 20.2 & \textbf{10.1} & 28.4 & \textbf{22.2} & 17.2 & \textbf{9.8} \\
\bottomrule
\end{tabular}
}
\caption{\textbf{Error composition, baseline
vs.\ \textsc{Parthenon}.} Same columns as Table~\ref{tab:baseline-summary}.
Each cell is the share of all criteria failing for a given cause
(judge-rationale mapping); cause rows sum to the \emph{Total error}
($=100-$ pooled accuracy). Bold marks the lower error rate in each
same-solver pair; API/Basic are unpaired. \textsc{Parthenon} lowers total
error at every solver and shrinks most causes. The ten classes are defined
in Appendix~\ref{app:error-examples}.}
\label{tab:residual-error-breakdown}
\end{table}

\paragraph{Difficulty varies by practice area; the kind of failure does
not.} The same Codex/GPT-5.5 error also splits by practice area. The
volume of failures varies widely across areas, tracking intrinsic matter
difficulty such as document load and rubric density rather than any
framework choice. The composition of those failures, however, is stable.
The same five named classes dominate the error in every area, so the
mix of error types is largely shared even where the difficulty is not.
A single set of cross-area controls can therefore target most of the
error, with per-practice knowledge layered on top.

               % 4.2 baseline error analysis
\subsection{Parthenon error reduction and comparison}
\label{subsec:patterns}

With the model and agent harness unchanged, this subsection traces
\emph{how} \textsc{Parthenon} lowers the error rate -- where it drops
(\S\ref{ssub:residual-drop}), what the agent does differently
(\S\ref{ssub:action-intensity}--\ref{ssub:action-mix}), whether output
length explains the gains
(\S\ref{ssub:deliverable-length}--\ref{ssub:trajectory-length}), and what
the lift costs (\S\ref{ssub:cost}). Each effect is read across the
four solver families: the headline tiers Codex/GPT-5.5 and Claude
Code/Sonnet 4.6, plus the lighter Codex/GPT-5.4-mini and Claude Code/Haiku
4.5.

\subsubsection{Error reduction by practice area}
\label{ssub:residual-drop}
\begin{figure}[H]
\centering
\includegraphics[width=\textwidth]{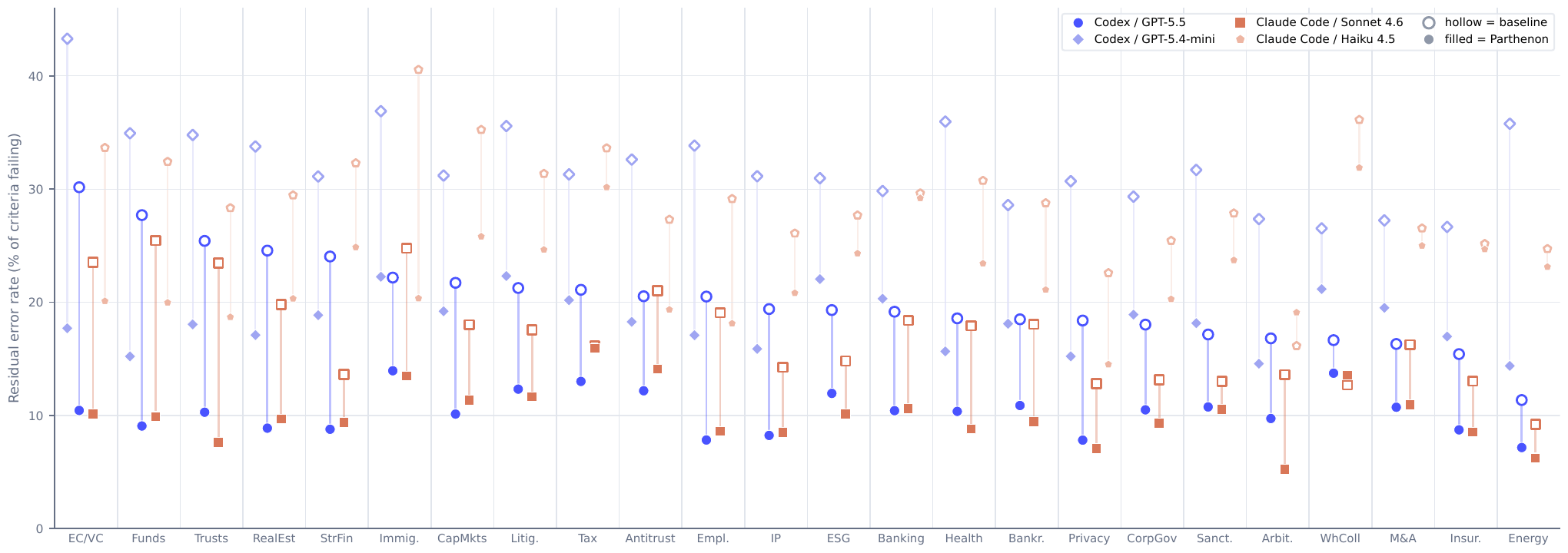}
\caption{\textbf{Error rate by practice area, baseline vs.\
\textsc{Parthenon}, four solver families.} Areas ordered hardest (left) to
easiest (right) by Codex/GPT-5.5 baseline error rate. One dumbbell per family per
area, grouped Codex (indigo, left) and Claude Code (terracotta, right): the
headline tiers Codex/GPT-5.5 (circles) and Claude Code/Sonnet 4.6 (squares) in
saturated color, the lighter tiers Codex/GPT-5.4-mini (faded diamonds) and
Claude Code/Haiku 4.5 (faded pentagons) on the outer offsets; hollow =
baseline, filled = \textsc{Parthenon}, connector = error drop. Every family
improves nearly every area (Codex $24/24$ at both tiers, Claude $23/24$),
most on the hardest, and the weaker base solvers (mini, Haiku) start higher
and drop further.}
\label{fig:domain-residual}
\end{figure}

The harness lowers the error rate in almost every practice area, and the
gains are largest where the baseline is weakest, so the cross-area spread
narrows (Figure~\ref{fig:domain-residual}). This holds across all four
solver families, including the much stronger Claude Code/Sonnet 4.6
baseline, which confirms the effect is not a low-baseline artifact. Because
the failure modes are shared across practice areas, a small set of
cross-practice controls -- source-coverage closure, numeric and date
reconciliation, deliverable validation, and authority grounding -- handles
most of the error, with practice-specific knowledge layered on top. This
is exactly the division \textsc{Parthenon} encodes as shared harness and
tools versus per-practice skills.

\subsubsection{Action intensity per matter}
\label{ssub:action-intensity}
\begin{figure}[H]
\centering
\includegraphics[width=\textwidth]{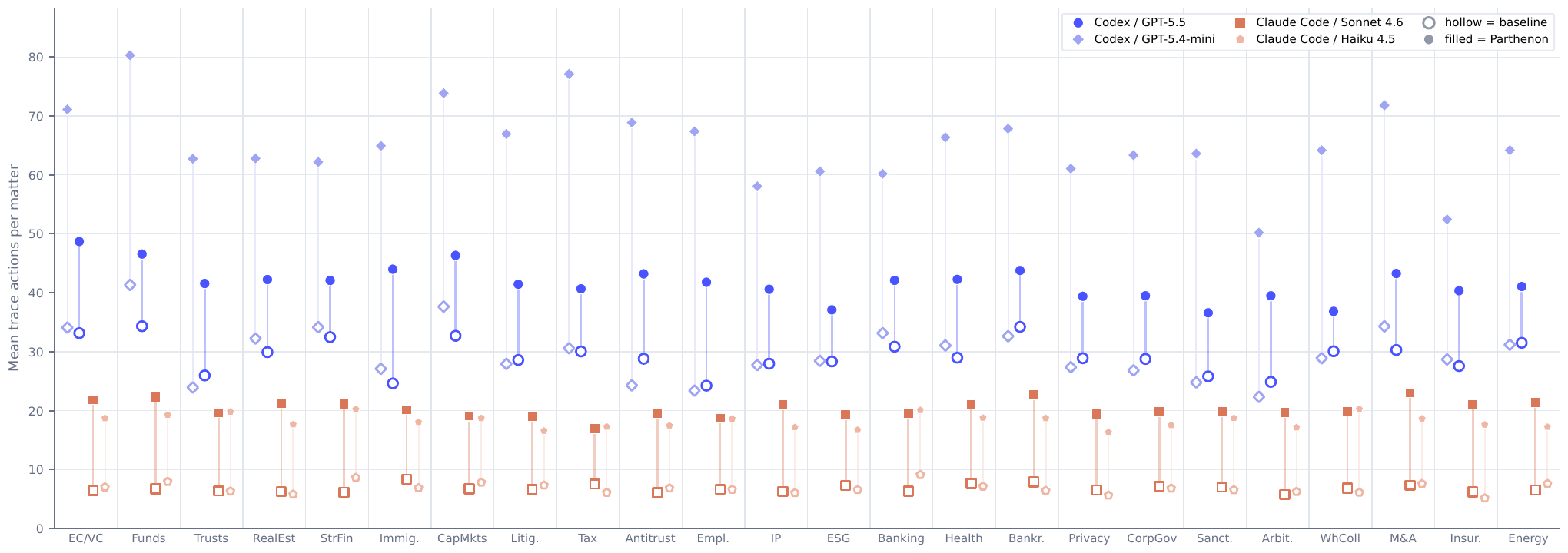}
\caption{\textbf{Actions per matter, $24$ practice areas, four solver
families.} Same area order and marker scheme as
Figure~\ref{fig:domain-residual}: headline tiers Codex/GPT-5.5 (circles) and
Claude Code/Sonnet 4.6 (squares) in saturated color, lighter tiers
Codex/GPT-5.4-mini (faded diamonds) and Claude Code/Haiku 4.5 (faded pentagons);
hollow = baseline, filled = \textsc{Parthenon}, connector = rise in actions. The
two trace formats count actions differently (an action is one tool call; Codex
uses many fine-grained shell commands -- the indigo band near the top --
whereas Claude Code issues fewer coarse tool calls -- the terracotta band
below), so read the within-family increase, not cross-family magnitude. Both
tiers of each family rise under \textsc{Parthenon}. Buckets in
Table~\ref{tab:action-mix}.}
\label{fig:where-lift-dumbbell}
\end{figure}

The framework adds more structured actions per matter
(Figure~\ref{fig:where-lift-dumbbell}). The added events are mandatory
inspections, retrievals, schema fills, and release-gate audits, and the
largest rises fall in the schema-heavy domains that the Skills layer most
expands. The two solver families count actions on different scales: Codex
issues many fine-grained shell commands, while Claude Code issues fewer
coarse tool calls, so the absolute counts are not comparable across
families. Read instead the within-family rise: for example, mean native
trace actions on the Claude Code/Sonnet 4.6 family rise from $6.8$ to
$20.6$ per matter.

\subsubsection{Action mix: a new tool/script bucket}
\label{ssub:action-mix}
The added actions form a new tool and script bucket rather than more
writing. Table~\ref{tab:action-mix} decomposes each cell's solver actions
into four behavioral buckets recovered from the renderable trace. The
baseline is a near-pure read trajectory dominated by \emph{read/inspect}
actions, with almost no \emph{tool/script} use. Under \textsc{Parthenon}
the new \emph{tool/script} bucket accounts for essentially the entire
increment: on Codex/GPT-5.5 it rises from $0.2$ to $16.5$ actions per
matter, made up of Skills-driven invocations of \texttt{summarize\_docs},
\texttt{deliverable\_search}, schema fillers, and audit utilities that the
baseline does not have. Over the same change, read/inspect edges down and
\emph{write/edit} barely moves, so the framework is not producing more
text but a structured tool layer. The pattern holds at every solver, the
tool/script bucket appearing where the baseline had none, while write and
edit stay a small fraction throughout. The matter-level shift is broadly
positive, a clear majority of paired tasks improving. The framework
changes \emph{what the solver does}, not how much text it emits.

\begin{table}[H]
\centering
\scriptsize
\setlength{\tabcolsep}{2.5pt}
\renewcommand{\arraystretch}{1.05}
\resizebox{\textwidth}{!}{%
\begin{tabular}{@{}l C{1.0cm}|C{1.0cm}||C{1.0cm}|C{1.0cm}||C{1.0cm}|C{1.0cm}||C{1.0cm}|C{1.0cm}@{}}
\toprule
Model & \multicolumn{2}{c||}{GPT-5.4-mini} & \multicolumn{2}{c||}{GPT-5.5} & \multicolumn{2}{c||}{Haiku 4.5} & \multicolumn{2}{c}{Sonnet 4.6} \\
Harness & Codex & \textsc{Parth.} & Codex & \textsc{Parth.} & Claude & \textsc{Parth.} & Claude & \textsc{Parth.} \\
\midrule
Read / inspect& 25.1 & \textbf{38.3} & \textbf{26.7} & 21.2 & 5.5 & \textbf{13.4} & 5.2 & \textbf{12.9} \\
Tool / script& 0.5 & \textbf{19.5} & 0.2 & \textbf{16.5} & 0.0 & \textbf{3.0} & 0.2 & \textbf{5.0} \\
Write / edit& 3.2 & \textbf{4.1} & 1.9 & \textbf{3.2} & 1.3 & \textbf{1.7} & 1.4 & \textbf{2.5} \\
Other& 1.1 & \textbf{3.6} & 0.5 & \textbf{0.9} & 0.0 & \textbf{0.1} & 0.0 & \textbf{0.2} \\
\midrule
Mean actions/matter& 30.0 & \textbf{65.4} & 29.3 & \textbf{41.8} & 6.9 & \textbf{18.2} & 6.8 & \textbf{20.6} \\
\bottomrule
\end{tabular}
}
\caption{\textbf{Action mix, baseline vs.\ \textsc{Parthenon}.} Mean number of
solver trace actions \emph{per matter} in each behavioral bucket, for the four
solver families with renderable traces; the bottom row is the mean total
actions per matter, and bold marks the larger value in each
baseline/\textsc{Parthenon} pair. Buckets: \emph{read/inspect}
(\texttt{rg}/\texttt{cat}/\texttt{sed}/\texttt{jq}/\texttt{wc}),
\emph{tool/script} (Skills interpreter calls), \emph{write/edit} (file
changes), \emph{other}. The framework adds a large \emph{tool/script} bucket at
every solver; per-area variation is small, so counts are pooled over the $24$
practice areas.}
\label{tab:action-mix}
\end{table}

\subsubsection{Deliverable length}
\label{ssub:deliverable-length}
\begin{figure}[H]
\centering
\includegraphics[width=\textwidth]{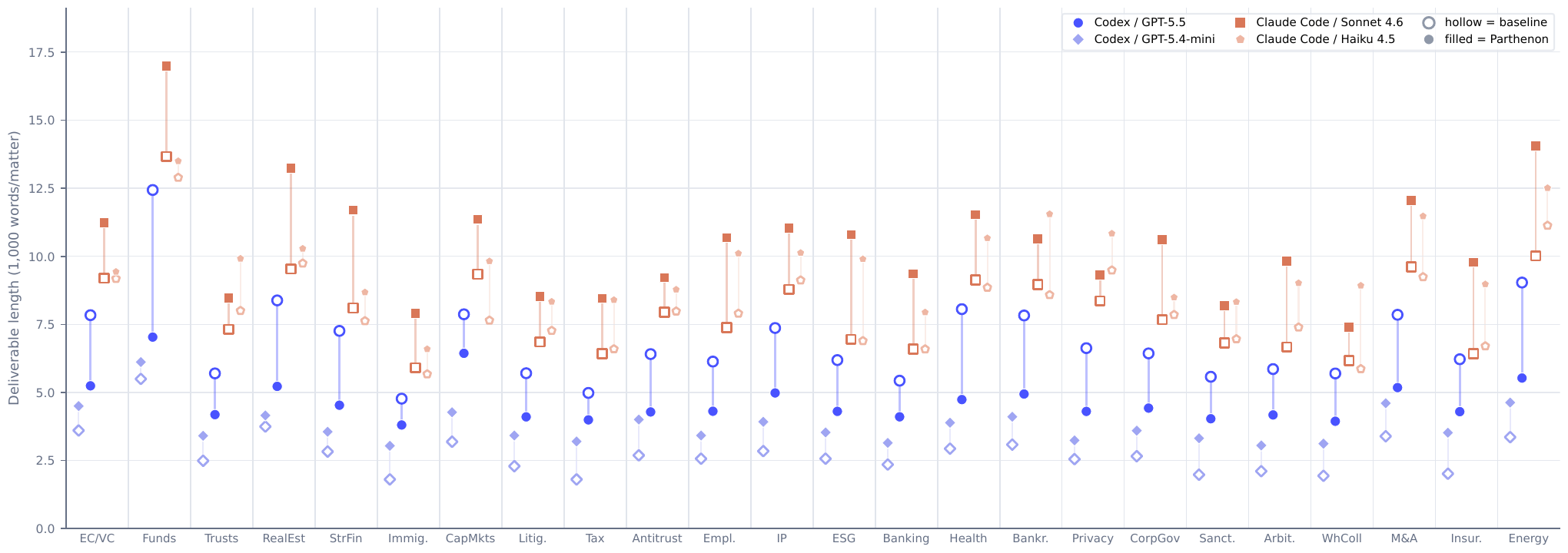}
\caption{\textbf{Deliverable length by practice area, baseline vs.\
\textsc{Parthenon}, four solver families ($1{,}000$ words/matter).} Same layout
and marker scheme as Figure~\ref{fig:where-lift-dumbbell}: Codex/GPT-5.5
(circles) and Sonnet 4.6 (squares) in saturated color, faded
Codex/GPT-5.4-mini (diamonds) and Haiku 4.5 (pentagons); hollow = baseline,
filled = \textsc{Parthenon}, connector = change. The direction flips by family:
GPT-5.5 deliverables \emph{shrink} while mini, Sonnet, and Haiku
\emph{grow}, yet accuracy rises for every family
(Table~\ref{tab:baseline-summary}), so length does not explain the gains.}
\label{fig:deliverable-length}
\end{figure}

Output length does not explain the gains. Deliverables shrink for some
solver families and grow for others, with the direction flipping by family,
yet accuracy rises for all four (Figure~\ref{fig:deliverable-length}). On
Codex/GPT-5.5 deliverables get shorter where accuracy rises most, while the
mini, Sonnet, and Haiku deliverables grow. Since more text accompanies both
larger and smaller accuracy gains, emitting more text is not the mechanism
behind the improvement.

\subsubsection{Agent-trajectory length}
\label{ssub:trajectory-length}
\begin{figure}[H]
\centering
\includegraphics[width=\textwidth]{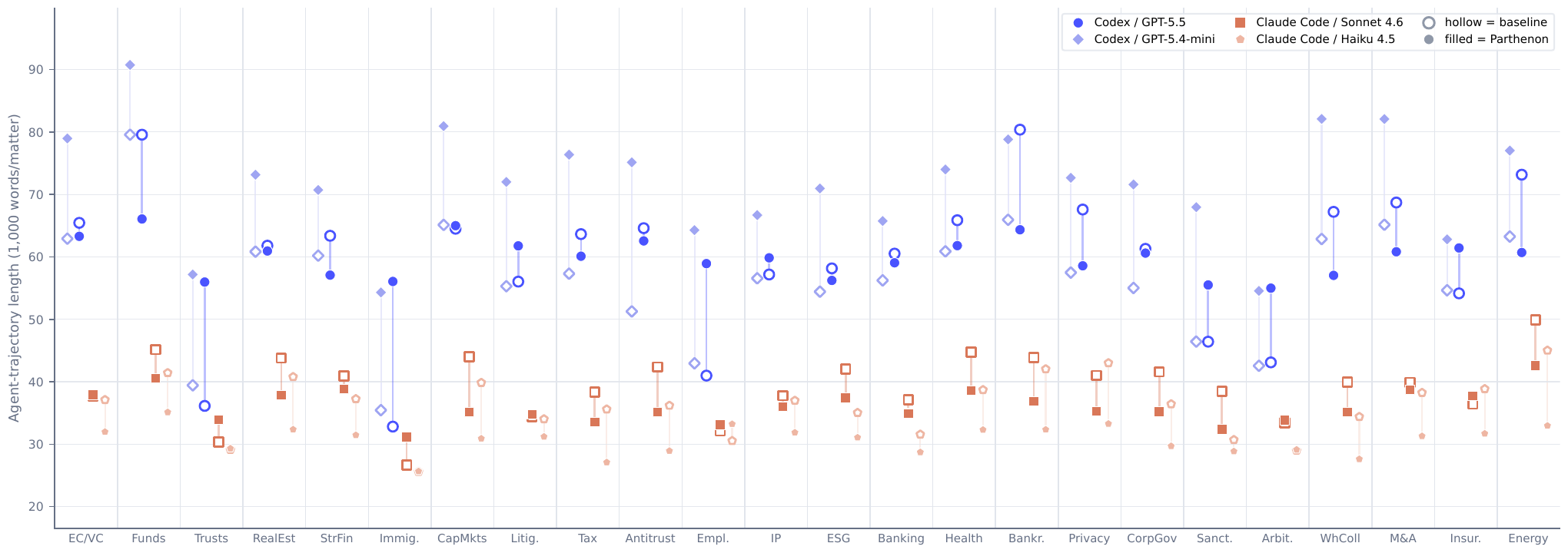}
\caption{\textbf{Agent-trajectory length by practice area, baseline vs.\
\textsc{Parthenon}, four solver families ($1{,}000$ words/matter).} Total agent
and tool text in the native trace; same layout and markers as
Figure~\ref{fig:deliverable-length}. The direction is again mixed -- the Codex
tiers (indigo) grow or hold while the Claude Code tiers (terracotta) shrink --
decoupled from the uniformly positive accuracy lift.}
\label{fig:main-result-length}
\end{figure}

The size of the lift tracks how hard the baseline found the matter, not how
much text the agent produces. Agent-trajectory length, the total agent and
tool text in the native trace, moves in mixed directions: the Codex tiers
grow or hold while the Claude Code tiers shrink
(Figure~\ref{fig:main-result-length}). This length is decoupled
from the uniformly positive accuracy lift. What the lift does track is
baseline difficulty, with the largest gains falling on the matters the
baseline handled worst and the matters it already passed gaining little.

\subsubsection{What the lift costs}
\label{ssub:cost}
\begin{figure}[H]
\centering
\includegraphics[width=\textwidth]{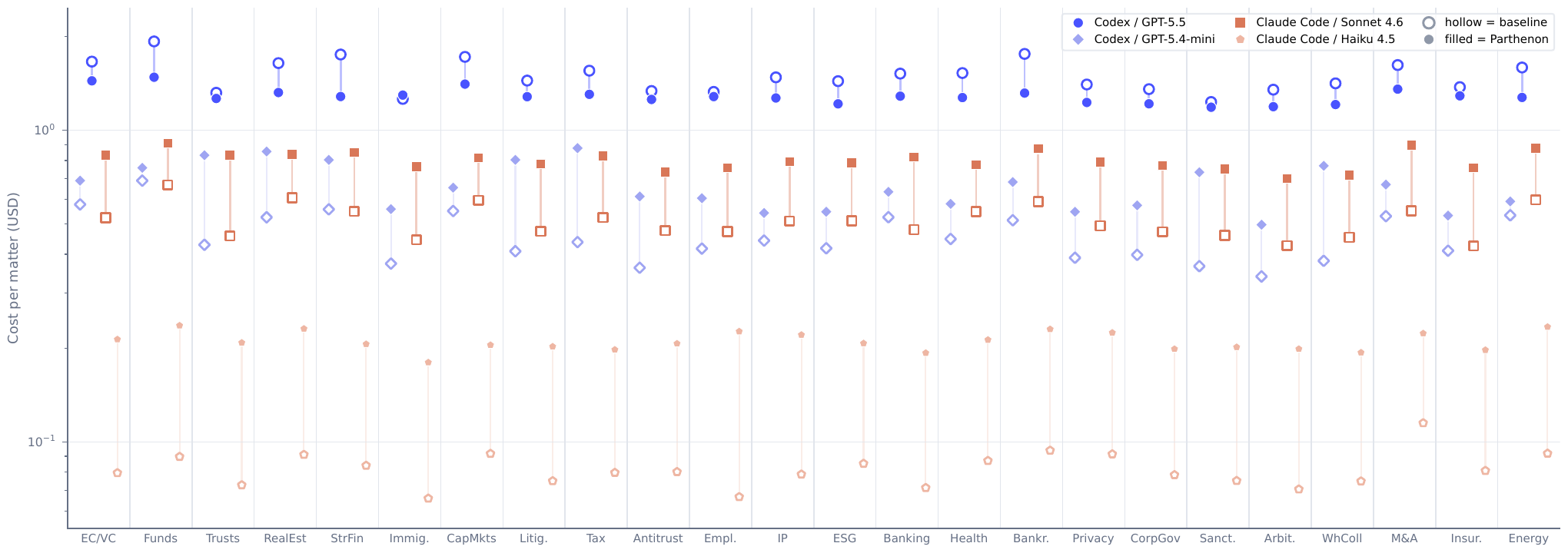}
\caption{\textbf{Cost per matter by practice area, baseline vs.\
\textsc{Parthenon}, four solver families.} Mean USD per matter
(\emph{log} scale) from logged token usage $\times$ public list prices,
cached-read-free; same cell layout and area order as
Figure~\ref{fig:where-lift-dumbbell}. Markers: Codex/GPT-5.5 (circles),
Codex/GPT-5.4-mini (faded diamonds), Claude Code/Sonnet 4.6 (squares),
Claude Code/Haiku 4.5 (faded pentagons); hollow = baseline, filled =
\textsc{Parthenon}. Costs span more than an order of magnitude across families;
\textsc{Parthenon}/GPT-5.5 is the only pair whose cost \emph{falls}
(\$1.51 $\to$ \$1.29), while the cheaper solvers spend more under the
framework. All cells are recomputed from logged tokens with cached reads
free, except the Codex/GPT-5.5 baseline, whose clean traces do not expose
billable tokens and whose cost is the original logged estimate. Judge and
evaluation tokens are excluded.}
\label{fig:cost-per-area}
\end{figure}

Whether the harness raises or lowers per-matter cost depends on the base solver.
Figure~\ref{fig:cost-per-area} prices each matter from logged token usage at
public list rates (cached reads free; judge and evaluation tokens excluded;
App.~\ref{app:cost-method}). One cell is an exception: the Codex/GPT-5.5
baseline traces do not expose billable token counts, so its cost is the
original full-cell logged estimate rather than a token recomputation. Read
against the accuracy of
Table~\ref{tab:baseline-summary}, the framework does not buy accuracy at a
fixed token premium. On the expensive, exploration-heavy GPT-5.5 the harness
is actually cheaper, at \$1.51\,$\to$\,\$1.29 per matter while accuracy rises
$79.8\to89.9\%$. The reason is that the same audit loop that adds tool calls
also yields shorter audited output, and on a solver whose output tokens
dominate the bill that reduction more than offsets the extra calls. The
cheaper solvers show the opposite sign, spending a little more under
\textsc{Parthenon}, but the surcharge is small relative to the accuracy it
buys. \textsc{Parthenon}/mini reaches $82.0\%$ for \$0.66, clearing the
GPT-5.5 baseline's $79.8\%$ at under half that baseline's cost.
\textsc{Parthenon}/Sonnet reaches the corpus-best $90.2\%$ for \$0.81. With
per-matter cost spanning more than an order of magnitude across families, the four pairings
offer a range of operating points: the cheapest acceptable draft from
\textsc{Parthenon}/mini, the highest accuracy from \textsc{Parthenon}/Sonnet,
and, uniquely at the top Codex tier, accuracy and cost improving together.

\paragraph{What remains.} Even with the strongest pairing
(\textsc{Parthenon}/Sonnet 4.6), $\sim 10\%$ of criteria still fail,
concentrated in recall and reasoning misses (missing facts,
numbers/dates) and rationales the scripted taxonomy does not yet name
(Table~\ref{tab:residual-error-breakdown}); the form- and grounding-class
errors are nearly gone. That remainder is the target of continued model
progress and the continual-learning loop (\S\ref{sec:learning}), not of
more workspace machinery. As a caveat, per-area accuracy is confounded
with task mix, rubric density, and document count, so these are
descriptive comparisons over a fixed task universe, not causal
attributions.

                 % 4.3 Parthenon error reduction & comparison
\subsection{Ablation study}
\label{subsec:ablations}

Three ablations probe \emph{where} the framework's gain comes from, and
where it does not. The first traces the self-improvement loop that builds
the harness; the second varies raw inference budget on a fixed harness;
the third adds a representative tool, cached document summaries. Read
together they point the same way: the dependable source of improvement is
retained, audited \emph{procedure}, rather than additional compute or any
single added tool.

\subsubsection{Harness optimization}
The self-improvement loop retains a transferable harness rather than the
output of a single favorable run. We run the loop (\S\ref{sec:learning}) on the
hardest ten tasks with two unrelated solver/learners -- Codex
(GPT-5.4-mini) and Claude Code (Haiku 4.5) -- each
starting from the same empty harness and using the same evaluation
panel and scoring procedure. The raw per-step candidates never settle, swinging by several
points between adjacent steps. A strict best-so-far gate that keeps only
audited gains converts this noise into a monotone, non-decreasing climb;
for Codex the gated accuracy rises from $45.3\%$ to $55.8\%$ over the run.
The central result is that the two solvers, despite different base models,
converge to within $0.4$\,pp of one another. The gated climb is stepwise
rather than smooth: durable gains arrive as discrete jumps when an audited
edit lands and then hold on a plateau -- the signature of adding a missing
control rather than tuning a model. What accumulates across steps is therefore
a transferable, inspectable harness, the product of selecting and retaining
audited edits rather than a single stochastic sample from either model.

\begin{figure}[H]
\centering
\includegraphics[width=\textwidth]{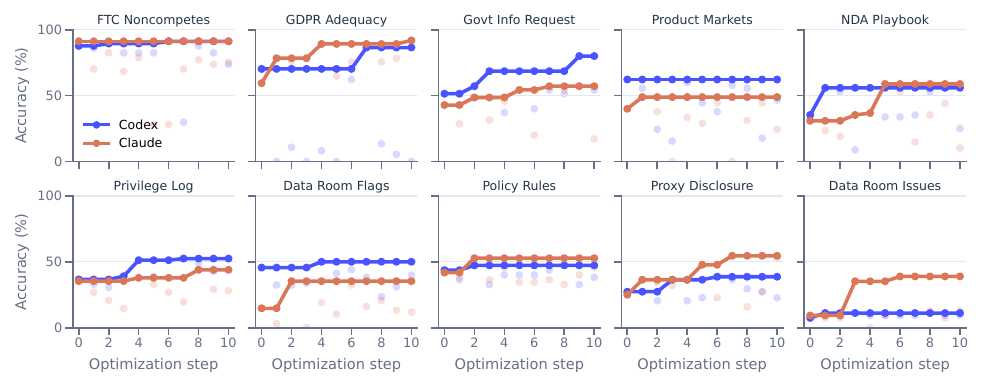}
\caption{\textbf{Per-task optimization trajectories.} One panel per
hard-$10$ task, ordered by final Codex accuracy. Solid lines: gated
best-so-far accuracy for Codex (GPT-5.4-mini) and Claude
Code (Haiku 4.5) as joint solver/learner; faint dots: raw
candidate pass rates. Strict best-so-far gate makes the frontier
monotone non-decreasing.}
\label{fig:optimization-curve}
\end{figure}

\subsubsection{Reasoning effort}
Raw inference budget helps on average but does so unreliably. On the same
ten tasks ($545$ criteria), increasing the coding-agent's effort setting
raises mean accuracy, yet neither monotonically nor consistently across
solvers. For Codex/GPT-5.5 the medium baseline improves at the
\texttt{high} setting and then regresses at the top \texttt{xhigh} setting.
Claude Code/Sonnet inverts this ordering: its medium baseline beats both
\texttt{low} and \texttt{high}, and only the \texttt{max} setting edges
ahead. Per-task winners scatter across settings rather than aligning on a
single level, and on several matters the extra budget amplifies an
unhelpful trajectory instead of correcting it. The two solvers often move
in opposite directions at the same setting, and at the highest setting a few
matters collapse outright -- accuracy falling by more than half from the
medium baseline -- as the extra reasoning entrenches a wrong approach. Set
against the monotone, transferable climb of the optimization loop
(Figure~\ref{fig:optimization-curve}), raw budget is the fragile lever and
audited procedure the dependable one. Effort is therefore a hyperparameter
worth tuning, but not a substitute for the procedural controls and
task-specific harness learning that move accuracy more dependably.

\begin{figure}[H]
\centering
\includegraphics[width=\textwidth]{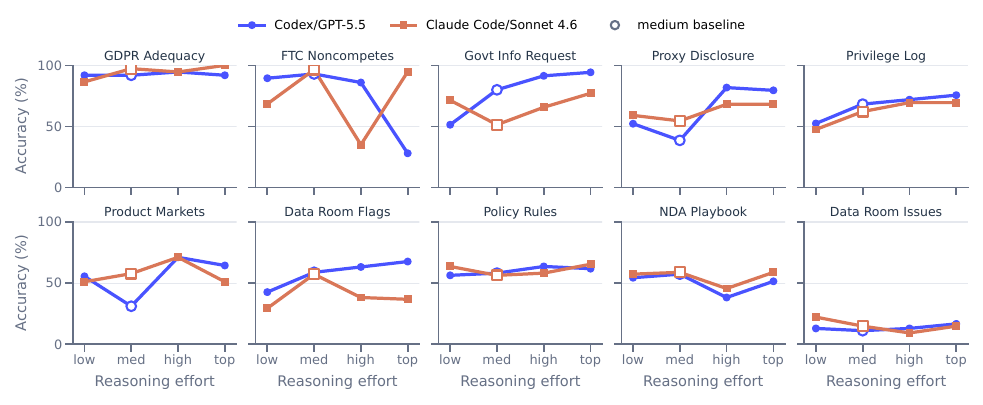}
\caption{\textbf{Reasoning-effort ablation.} Same hard-$10$
tasks as Figure~\ref{fig:optimization-curve}, run with Codex
(GPT-5.5) and Claude Code (Sonnet 4.6) while
varying the coding-agent CLI effort setting. Hollow markers denote
the medium-effort baseline. The \texttt{top} tick is \texttt{xhigh}
for Codex and \texttt{max} for Claude Code; other cells are matched
low/high/top reruns. Panels are ordered by best observed accuracy
across effort levels.}
\label{fig:reasoning-effort-ablation}
\end{figure}

\subsubsection{Document-summary tool}
The long-input gap reflects retrieval discipline, not the absence of a
compact summary. A cached per-document summary is the obvious remedy for
matters with many sources, yet enabling it does not pay off. Per-task
deltas center on zero and tilt slightly negative, so the cache modestly
hurts short matters. The reason is that the real handicap is input length
itself: accuracy declines steadily as the number of source documents grows,
and the cache leaves that decline essentially untouched. If a compact
overview were the fix, its benefit should grow with corpus size; instead it
shows no trend with input length and gains and losses roughly cancel, so the
tool fails exactly where long inputs hurt most. The sharpest
regressions all involve row-level fact work (\texttt{identify},
\texttt{extract}, \texttt{compare}, \texttt{review}, \texttt{assess}),
where the solver consumes the summary as compressed source text rather than
using it as an index back into the originals. The lift therefore comes from
procedural controls that enforce disciplined retrieval and verification,
not from shorter or orienting context.

\begin{figure}[H]
\centering
\includegraphics[width=\textwidth]{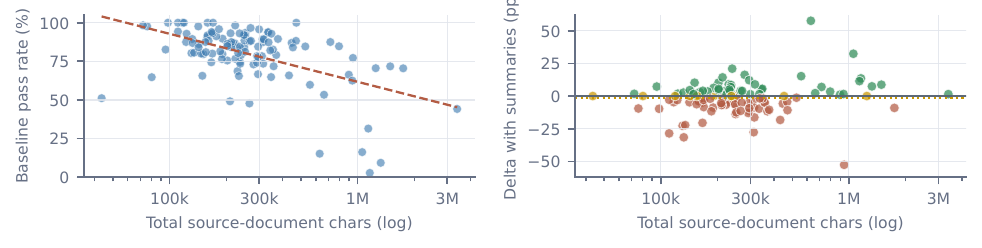}
\caption{\textbf{Document-summary tool ablation.} Both panels use
the same paired GPT-5.4-mini Codex tasks. (A) Baseline pass
rate vs.\ total source-document chars (log); pass rate declines
as inputs grow. (B) Per-task delta from enabling the cached
document-summary tool, centered near zero; the tool does not
close the long-input gap.}
\label{fig:doc-summary-tool-ablation}
\end{figure}
            % 4.4 targeted ablations
\subsection{Human-lawyer reference: accuracy, time, and cost}
\label{subsec:human-comparison}

The deployment target is to accelerate a supervising lawyer, not to replace one.
Harvey LAB has no controlled human-lawyer run on the $1{,}251$ matters.
Rather than invent a human accuracy score, we treat the lawyer as the
\emph{release} standard that the benchmark approximates. A work product is
acceptable only when every material fact, authority, deadline, number, and
deliverable requirement survives review, an all-or-nothing standard we
denote $100\%$. Two readings of Table~\ref{tab:human-reference} pull in
opposite directions. On accuracy, the system is still a draft assistant.
Even \textsc{Parthenon}'s strongest configuration (Claude Code/Sonnet 4.6)
passes $90.2\%$ of criteria, yet it clears \emph{every} criterion on only
$150$ matters, leaving $1{,}101$ that a lawyer must still finish. On time
and cost, the comparison already favors the system, though the two sides are
measured differently. Per-matter cost is computed from logged tokens at
\$0.81, thousands of times below the \$$4{,}399$ blended human estimate.
Agent wall-clock was not logged, so the time gap is an order-of-magnitude
estimate rather than a measurement: against a human's $12.6$\,h blended
figure ($3$--$20$\,h by complexity), a per-complexity bucket estimate of
roughly ten minutes per matter implies a speedup on the order of $70\times$.
Both gaps precede any review. The implication is a
change of role rather than a replacement. The framework converts the
expensive part of a matter from drafting from scratch into reviewing a
source-grounded, audit-flagged first draft, so the lawyer's hours go to
residual legal judgment rather than document assembly. The accuracy gap
says it is not autonomous; the time-and-cost gap says it is already useful.

\begin{table}[H]
\centering
\scriptsize
\setlength{\tabcolsep}{3pt}
\renewcommand{\arraystretch}{1.02}
\resizebox{\textwidth}{!}{%
\begin{tabular}{@{}lrrrrrr@{}}
\toprule
& \multicolumn{2}{c}{\textbf{Accuracy (\%)}} & \multicolumn{2}{c}{\textbf{Time / matter}} & \multicolumn{2}{c}{\textbf{Cost / matter (USD)}} \\
\cmidrule(lr){2-3} \cmidrule(lr){4-5} \cmidrule(lr){6-7}
\textbf{Practice area} & Human & \textsc{Parthenon} & Human & \textsc{Parthenon} & Human & \textsc{Parthenon} \\
\midrule
International Arbitration \& Dispute Resolution & 100 & 94.8 & 20\,h & 820\,s & 6{,}980 & 0.70 \\
Energy \& Natural Resources & 100 & 93.8 & 10\,h & 540\,s & 3{,}490 & 0.88 \\
Data Privacy \& Cybersecurity & 100 & 92.9 & 3\,h & 290\,s & 1{,}047 & 0.79 \\
Trusts, Estates \& Private Client & 100 & 92.4 & 20\,h & 740\,s & 6{,}980 & 0.83 \\
Intellectual Property & 100 & 91.5 & 10\,h & 580\,s & 3{,}490 & 0.79 \\
Insurance & 100 & 91.5 & 10\,h & 500\,s & 3{,}490 & 0.76 \\
Employment \& Labor & 100 & 91.4 & 3\,h & 310\,s & 1{,}047 & 0.76 \\
Healthcare \& Life Sciences & 100 & 91.2 & 10\,h & 560\,s & 3{,}490 & 0.77 \\
Corporate Governance & 100 & 90.7 & 10\,h & 600\,s & 3{,}490 & 0.77 \\
Structured Finance \& Securitization & 100 & 90.6 & 20\,h & 820\,s & 6{,}980 & 0.85 \\
Bankruptcy \& Restructuring & 100 & 90.6 & 20\,h & 800\,s & 6{,}980 & 0.87 \\
Real Estate & 100 & 90.3 & 20\,h & 770\,s & 6{,}980 & 0.84 \\
Funds \& Asset Management & 100 & 90.1 & 20\,h & 880\,s & 6{,}980 & 0.91 \\
Emerging Companies \& Venture Capital & 100 & 89.9 & 20\,h & 810\,s & 6{,}980 & 0.83 \\
Environmental \& ESG & 100 & 89.9 & 10\,h & 530\,s & 3{,}490 & 0.79 \\
International Trade \& Sanctions & 100 & 89.5 & 3\,h & 300\,s & 1{,}047 & 0.75 \\
Banking \& Finance & 100 & 89.4 & 10\,h & 580\,s & 3{,}490 & 0.82 \\
Corporate M\&A & 100 & 89.1 & 20\,h & 860\,s & 6{,}980 & 0.90 \\
Capital Markets & 100 & 88.6 & 20\,h & 760\,s & 6{,}980 & 0.81 \\
Litigation \& Dispute Resolution & 100 & 88.4 & 10\,h & 620\,s & 3{,}490 & 0.78 \\
Immigration & 100 & 86.5 & 3\,h & 280\,s & 1{,}047 & 0.76 \\
White Collar Defense \& Investigations & 100 & 86.5 & 10\,h & 510\,s & 3{,}490 & 0.72 \\
Antitrust \& Competition & 100 & 85.9 & 10\,h & 590\,s & 3{,}490 & 0.73 \\
Tax & 100 & 84.1 & 10\,h & 540\,s & 3{,}490 & 0.83 \\
\midrule
\textbf{Mean} & \textbf{100} & \textbf{90.2} & \textbf{12.6\,h} & \textbf{626\,s} & \textbf{4{,}399} & \textbf{0.81} \\
\bottomrule
\end{tabular}
}
\caption{Per-practice-area comparison against the release standard ($24$ Harvey LAB
domains, sorted by AI accuracy). \emph{Accuracy:} AI is the measured
\textsc{Parthenon} (Claude Code/Sonnet 4.6, our strongest
configuration) criterion accuracy; human accuracy is the release
standard LAB approximates. \emph{Time:} human times are
matter-complexity bucket estimates (3\,h screen / 10\,h diligence
/ 20\,h complex); AI times are matter-complexity bucket estimates.
\emph{Cost:} human at \$349/h \citep{clio2026rates}; AI cost is the
measured per-matter cost from logged tokens (cached-read-free).}
\label{tab:human-reference}
\end{table}

\section{Conclusion}
\label{sec:conclusion}

Stronger base models still fail legal work for a reason that is largely
procedural rather than parametric: across models and practice areas, agents
fail in the same professional categories -- quantitative detail, missing
facts, deliverable form, rule application, and source grounding -- because
the workspace enforces no contract for when a matter is complete.
\textsc{Parthenon} closes this gap without retraining, wrapping the solver
in an auditable legal harness: deterministic tools that enforce a matter's
invariants, procedural skills that encode how to work a matter class, and a
self-evolving loop that turns scored failures into reviewable harness diffs
under strict anti-leakage. With the base model and agent runtime fixed,
the harness delivers a lift comparable to a model upgrade at every solver tier
and transfers across unrelated base models, so the gain is a property of the
harness rather than of any one model.
The system is not an autonomous lawyer. Against an all-or-nothing release
bar it still clears every criterion on only a minority of matters, so its
role is to accelerate a supervising lawyer: it converts drafting from scratch into
reviewing a source-grounded, audit-flagged draft and leaves residual legal
judgment to the attorney. More broadly, when reliability is bounded by
procedure rather than raw capability, an external harness that a human can
inspect and a system can improve from its own failures may carry further
than the next model.
 % 4.5 human-lawyer reference

\bibliographystyle{plainnat}
\bibliography{references}

\appendix

\begin{center}
  {\Large\bfseries Appendix}
\end{center}
\vspace{0.5\baselineskip}

\section{Evaluation Protocol and Measurement}
\label{app:data-pack}

This appendix records the measurement details that the main text uses but
does not spell out: the corpus and scoring definitions, how raw traces are
turned into the action buckets of \S\ref{ssub:action-intensity}, and how the
cost and time figures of \S\ref{ssub:cost} and \S\ref{subsec:human-comparison}
are computed. It is deliberately not a per-task data dump: the full scored
outputs, native transcripts, and derived metrics are released with the
project data pack (App.~\ref{app:data-availability}).

\subsection{Corpus, cells, and metrics}
\label{app:protocol}

\paragraph{Benchmark universe.} The main results use the full Harvey
LAB corpus, not the smaller hard-task optimization subset:
$1{,}251$ matters, $24$ practice areas,
median $57$ criteria per matter (range $23$--$194$). Source-file
counts are reconstructed from the task corpus, which gives
median $7$ source files per matter.
The hard-$10$ optimization experiment (App.~\ref{app:optimization-protocol})
is a separate harness-learning panel and is not pooled into the
full-corpus tables.

\paragraph{Scoring formulas.} For scored run set $R$, let task $i$
have $c_i$ criteria and $p_i$ passed criteria. The paper reports
\[
\mathrm{PooledPass}=\frac{\sum_i p_i}{\sum_i c_i},\qquad
\mathrm{AllPass}=\frac{1}{|R|}\sum_{i\in R}\mathbf{1}[p_i=c_i].
\]
Pooled pass is criterion-weighted; all-pass is matter-level and
requires every criterion in a matter to pass. The two diverge sharply
because all-pass is a conjunction: a single missed criterion fails the
matter, so the metric is far more sensitive to errors than the
pooled average.

\begin{table}[H]
\centering
\scriptsize
\setlength{\tabcolsep}{4pt}
\renewcommand{\arraystretch}{1.08}
\begin{tabular}{@{}lrrr@{}}
\toprule
Cell & Runs & All-pass & Pooled pass \\
\midrule
Direct API / GPT-5.4-mini & 1{,}251 & 5 (0.40\%) & 56.5\% \\
Basic / GPT-5.4-mini & 1{,}251 & 2 (0.16\%) & 60.7\% \\
Codex / GPT-5.4-mini & 1{,}251 & 14 (1.12\%) & 68.2\% \\
Claude Code / Haiku 4.5 & 1{,}251 & 14 (1.12\%) & 71.6\% \\
Claude Code / Sonnet 4.6 & 1{,}251 & 148 (11.83\%) & 82.8\% \\
Codex / GPT-5.5 & 1{,}251 & 47 (3.76\%) & 79.8\% \\
\textsc{Parthenon} / GPT-5.4-mini & 1{,}251 & 42 (3.36\%) & 82.0\% \\
\textsc{Parthenon} / GPT-5.5 & 1{,}251 & 137 (10.95\%) & 89.9\% \\
\textsc{Parthenon} / Haiku 4.5 & 1{,}251 & 38 (3.04\%) & 77.8\% \\
\textsc{Parthenon} / Sonnet 4.6 & 1{,}251 & 150 (11.99\%) & 90.2\% \\
\bottomrule
\end{tabular}

\caption{Run inventory used by Table~\ref{tab:baseline-summary}. Listed cells cover the full
$1{,}251$-matter corpus.}
\label{tab:appendix-cell-inventory}
\end{table}

\paragraph{Basic legal-native harness.}
\label{app:harness-note}
The Basic column is the benchmark-provided native runner, not
Harvey's proprietary product environment. It is the open-source
\citeauthor{harvey2026labharness} harness contract released with
Harvey LAB \citep{harvey2026labharness}: tasks are filesystem workspaces, the
harness loads prompts, state, model adapters, and workspace tools,
and separate evaluation/reporting steps score outputs against rubric
criteria \citep{harvey2026labarchitecture}. We call it
\emph{Basic} to distinguish it from external workspace
agents such as Codex and Claude Code.

\paragraph{GPT-5.5 error extraction.}
The baseline error analysis (\S\ref{subsec:trajectory}) and the lift
decomposition (\S\ref{subsec:patterns}) use the recursive Codex/GPT-5.5
cell over the full corpus: $1{,}251$ scored runs, $59{,}819$ passed and
$15{,}171$ failed criteria. The failure taxonomy is assigned over those
failed criteria, and the per-area composition figures are computed from
the same assignments.

\subsection{From traces to action buckets}
\label{app:action-buckets}

The per-matter action counts in \S\ref{ssub:action-intensity}--%
\ref{ssub:action-mix} are recovered from the native execution traces and
sorted into four buckets -- \emph{read/inspect}, \emph{tool/script},
\emph{write/edit}, and \emph{other} -- by a deterministic classifier, with
unrecognized actions falling through to \emph{other}. The two solver
families expose different trace formats, so the classifier is format-aware.

\paragraph{Codex (shell grammar).} Codex emits fine-grained shell commands.
Each executed command is bucketed by its leading verb: file readers and
search utilities (\texttt{rg}, \texttt{grep}, \texttt{cat}, \texttt{sed},
\texttt{jq}, \texttt{wc}, \texttt{head}, \texttt{ls}, \texttt{find}) count
as \emph{read/inspect}; interpreters and build commands (\texttt{python},
\texttt{node}, \texttt{bash}, \texttt{pytest}, \texttt{latexmk}) as
\emph{tool/script}; file mutations (\texttt{apply\_patch}, \texttt{tee},
\texttt{mv}, \texttt{cp}, \texttt{touch}, \texttt{mkdir}) and structured
file-change events as \emph{write/edit}.

\paragraph{Claude Code (named-tool grammar).} Claude Code emits fewer,
coarser named tool calls. These are bucketed by tool name: \texttt{Read},
\texttt{Glob}, \texttt{Grep}, \texttt{LS} as \emph{read/inspect};
\texttt{Write}, \texttt{Edit}, \texttt{MultiEdit} as \emph{write/edit}; and
shell- or interpreter-style calls (\texttt{Bash}, \texttt{Skill},
\texttt{Agent}) as \emph{tool/script}, with embedded shell commands routed
through the same verb rules as above. Because one Codex shell command and one
Claude Code tool call are not the same unit of work, absolute counts are not
comparable across families; the main text reads only the within-family change.

\subsection{Cost and time accounting}
\label{app:cost-method}

\paragraph{Cost.} Per-matter cost is logged token usage priced at public
list rates (Table~\ref{tab:list-prices}). Cached-read tokens are treated as
free, and judge and evaluation tokens are excluded, so the figures reflect
solver inference only. All cells are recomputed from logged token counts
with one exception: the Codex/GPT-5.5 \emph{baseline} traces do not expose
billable token fields, so its per-matter cost (\$1.51) is the original
full-cell logged estimate rather than a recomputation. The corresponding
\textsc{Parthenon}/GPT-5.5 cell does expose tokens and is recomputed (\$1.29);
the ``cost falls under the harness'' comparison for this tier therefore
weighs a recomputed value against a logged estimate and should be read as
order-of-magnitude rather than exact.

\begin{table}[H]
\centering
\scriptsize
\setlength{\tabcolsep}{6pt}
\renewcommand{\arraystretch}{1.08}
\begin{tabular}{@{}lrrr@{}}
\toprule
Model & Input & Cached read & Output \\
\midrule
GPT-5.4-mini (OpenAI)   & 0.75 & 0.075 & 4.50 \\
GPT-5.5 (OpenAI)        & 5.00 & 0.50  & 30.00 \\
Haiku 4.5 (Anthropic)   & 1.00 & 0.10  & 5.00 \\
Sonnet 4.6 (Anthropic)  & 3.00 & 0.30  & 15.00 \\
\bottomrule
\end{tabular}
\caption{Public list prices used for the cost figure, in USD per $1$M
tokens (accessed 2026-06-02). Cached reads are treated as free in our
accounting. GPT-5.5 long-context requests above the documented threshold
are billed at higher rates; the recomputed cells apply the standard rates
to logged token fields.}
\label{tab:list-prices}
\end{table}

\paragraph{Time.}
\label{app:human-assumptions}
We did not log wall-clock time for the agent runs, so the per-matter
\emph{time} columns in Table~\ref{tab:human-reference} are
matter-complexity bucket estimates, not measured run durations: human work
is bucketed at $3$/$10$/$20$\,h by matter complexity, and the AI side uses a
matched per-complexity estimate. The time comparison (and the ``$\sim$$70\times$
faster'' statement) is therefore an order-of-magnitude claim about the scale
of the gap, whereas the cost comparison is grounded in measured tokens. The
human cost column applies a single blended rate of \$349/h
\citep{clio2026rates} to the same complexity buckets; the human accuracy
column is the all-criteria release standard that LAB approximates, not
a controlled human run.

\subsection{Information boundaries}
\label{app:info-boundaries}

\begin{table}[H]
\centering
\scriptsize
\setlength{\tabcolsep}{4pt}
\renewcommand{\arraystretch}{1.08}
\begin{tabularx}{\textwidth}{@{}L{2.2cm}L{3.15cm}Y@{}}
\toprule
Role / stage & Receives & Excluded by design \\
\midrule
Solver & Task brief, source files, deliverable spec, current
harness, legal tools and skills. & Criterion ids, criterion titles,
match criteria, judge rationales, failed-criterion text. \\
Evaluator & Finished work product, task brief, deliverable spec,
hidden rubric. & Solver context during drafting; no feedback channel
back into the active run. \\
Learner & Redacted traces, aggregate pass/fail signals,
de-identified failure reasons, current harness repository. &
Task ids as reusable content, source facts, client identities,
rubric phrases, answer keys. \\
Acceptance gate & Proposed harness diff, static checks, scored outcome
delta, leakage checklist. & Automatic promotion of task-specific
conclusions; any edit that only helps the exposing matter. \\
\bottomrule
\end{tabularx}
\caption{Anti-leakage contract used by the continual-learning loop.
The main full-corpus \textsc{Parthenon} result uses the optimized
harness with bounded revision disabled; the hard-$10$ ledger
(App.~\ref{app:optimization-protocol}) measures harness learning itself.}
\label{tab:info-boundaries}
\end{table}

\section{Framework Internals, Failure Taxonomy, and Ablation Protocols}
\label{app:error-examples}

\begingroup
\sloppy

This appendix expands the parts of the framework that the main text states
compactly: the concrete Tools surface (\S\ref{app:tools-detail}), the
structure of a Skill (\S\ref{app:skill-anatomy}), the failure taxonomy and
how rationales are mapped to it (\S\ref{app:taxonomy}), and the protocols
behind the ablations of \S\ref{subsec:ablations}
(\S\ref{app:optimization-protocol}--\ref{app:ablation-protocols}). It reports
protocols and representative cases rather than per-run results, which are
released in full with the data pack (\S\ref{app:data-availability}).

\subsection{The Tools surface in detail}
\label{app:tools-detail}

Table~\ref{tab:tools-catalogue-detail} groups the Tools layer into four
lifecycle stages; Table~\ref{tab:tools-detail} lists the $14$ agent-callable
tools individually. Each is a deterministic operation registered in a single
tool registry, so adding a capability is one reviewable diff. Several
single-file probes are prompt-level capabilities that drive standard search
utilities (\texttt{rg}/\texttt{sed}) against the canonicalized source text
rather than carrying bespoke parsers.

\begin{table}[H]
\centering
\scriptsize
\setlength{\tabcolsep}{4pt}
\renewcommand{\arraystretch}{1.1}
\begin{tabularx}{\textwidth}{@{}L{2.6cm}Y@{}}
\toprule
\textbf{Tool} & \textbf{Function} \\
\midrule
\multicolumn{2}{@{}l}{\emph{Pre-draft: canonicalize sources and retrieve knowledge}}\\
\addlinespace[1pt]
\texttt{summarize\_docs} & Build structured, line-numbered briefs (entities, operative terms, quantitative and date anchors) for every source file. \\
\texttt{authority} & Retrieve statutes, regulations, and case law from a local catalog (token-overlap ranking) with web fallback. \\
\texttt{skeleton} & Infer the deliverable type and return its required-section skeleton and issue/authority preflight. \\
\midrule
\multicolumn{2}{@{}l}{\emph{Single-file probes: targeted inspection of one source}}\\
\addlinespace[1pt]
\texttt{xlsx\_inspect} & Query spreadsheet extracts by label/value with scale-hint detection. \\
\texttt{eml\_inspect} & Query email extracts for headers (from/to/date/subject) and body sections. \\
\texttt{docx\_inspect} & Navigate Word extracts by article/section heading and defined terms. \\
\texttt{pptx\_inspect} & Query slide-deck extracts (one line per slide). \\
\texttt{citation\_scan} & Sweep a file for legal citations (U.S.C./C.F.R./F.R.C.P./case/\S). \\
\texttt{number\_scan} & Find numeric values (currency, percentages, bps, multiples, day-spans) in a file. \\
\texttt{date\_scan} & Find dates and deadline windows (``within $N$ days'', explicit and ISO dates). \\
\midrule
\multicolumn{2}{@{}l}{\emph{Post-draft: audit and emit the work product}}\\
\addlinespace[1pt]
\texttt{reconcile\_numbers} & Match every figure in the draft against the source corpus and flag ungrounded values (unit mismatch, fabricated, unsourced). \\
\texttt{reconcile\_deadlines} & Match every date and window claim in the draft against source dates and flag unsourced claims. \\
\texttt{audit\_all} & One-shot dispatcher: run the reconcilers and the skeleton check in parallel and return one merged release report. \\
\texttt{build\_deliverable} & Canonical generator: convert Markdown to \texttt{.docx} or tabular data to single/multi-sheet \texttt{.xlsx}. \\
\bottomrule
\end{tabularx}
\caption{The $14$ deterministic, agent-callable tools, grouped by matter
lifecycle stage. Five carry a mandatory self-audit step ordered into the
release gate (number then date grounding, skeleton compliance, then citation
accuracy); a trailing reasoning/completeness pass is always last.}
\label{tab:tools-detail}
\end{table}

\subsection{Anatomy of a skill}
\label{app:skill-anatomy}

Each of the $1{,}251$ skills is a versioned Markdown file routed to a matter
by its task identifier, with a short YAML header (\texttt{name},
\texttt{task\_id}, \texttt{description}, and the agent roles it activates for)
followed by the seven-part scaffold of the Skills layer. The
scaffold is rubric-blind by construction: it names \emph{procedure}, not
answers. Before promotion, a separate rubric-blind audit -- given only the
task title and the skill text -- rewrites any residual task-specific content
into generic procedure, so deployed skills carry the procedural sections and
delegate leakage control to that audit. Table~\ref{tab:skill-anatomy} shows
the scaffold instantiated, in abstracted form, for a litigation
case-assessment memorandum.

\begin{table}[H]
\centering
\scriptsize
\setlength{\tabcolsep}{4pt}
\renewcommand{\arraystretch}{1.12}
\begin{tabularx}{\textwidth}{@{}L{3.0cm}Y@{}}
\toprule
\textbf{Scaffold part} & \textbf{Generic content (example: case-assessment memo)} \\
\midrule
1.\ Subject-matter triage & Identify the operative record: which agreement governs, which causes of action are in scope, how many distinct claims/periods/damages theories to analyze. \\
2.\ Failure modes corrected & Patterns to avoid: analyzing claims in the abstract, accepting the demand as given, omitting coverage/forum, ending in summary rather than a disposition. \\
3.\ Legal frameworks / conventions & Doctrinal frame for the deliverable type: formation/breach/causation/damages, limitations and tolling, punitive eligibility, insurance coverage -- each cited to controlling authority by name. \\
4.\ Analytical scaffolds & The analytic sequence: map each claim to contract language and the timeline; per claim run standard\,$\to$\,supporting\,$\to$\,adverse\,$\to$\,risk\,$\to$\,consequence; decompose damages and test each. \\
5.\ Relationships (vertical / temporal) & How the analysis tracks time and hierarchy: notice/breach/cure/termination chronology; agreement vs.\ later conduct; shared-event interactions among claims. \\
6.\ Output structure conventions & The required form: subject line, executive summary with bottom-line view, background, claim-by-claim analysis, damages range, defenses, coverage, forum, disposition and next steps. \\
\midrule
Anti-leakage checklist & Forbidden in the body: task ids, rubric phrases, client identities, deal amounts, dates, or quotations from the matter that exposed the failure. \\
\bottomrule
\end{tabularx}
\caption{The seven-part skill scaffold, abstracted to procedure. A skill
encodes how to work a matter class, never an answer; the example is rendered
generically to show shape, not memorized content.}
\label{tab:skill-anatomy}
\end{table}

\subsection{Error taxonomy and how rationales are mapped}
\label{app:taxonomy}

Table~\ref{tab:error-taxonomy-controls} gives the operational meaning of the
ten error classes used in \S\ref{subsec:trajectory} and names the
\textsc{Parthenon} surface expected to absorb each. The mapping from a judge
rationale to a class is deterministic, not an LLM call: a priority-ordered
keyword/regex matcher reads the concatenated criterion title and rationale
and assigns the first class whose pattern fires (quantitative/temporal terms
first, then grounding, remediation, materiality, cross-document, deliverable
form, legal-rule, coverage, and missing-fact patterns). Rationales that match
no pattern fall through to \emph{Other}, which is therefore a residual bucket
to be split by future taxonomy work, not a named failure mode. Counts are
the criterion failures behind the Codex/GPT-5.5 column of
Table~\ref{tab:residual-error-breakdown}.

\begin{table}[H]
\centering
\scriptsize
\setlength{\tabcolsep}{3pt}
\renewcommand{\arraystretch}{1.08}
\begin{tabularx}{\textwidth}{@{}L{2.35cm}rrYY@{}}
\toprule
Class & Count & Share & Failure unit & Control surface \\
\midrule
Numbers/dates & 2{,}508 & 16.5\% & Missing amount, date,
threshold, formula, or deadline. & Number/date ledger; deadline
window lookup; final reconciliation gate. \\
Missing facts/entities & 3{,}497 & 23.1\% & Party, exhibit,
attribute, transaction fact, or entity status omitted. & Source
inventory; entity/fact ledger; coverage manifest. \\
Deliverable/format & 1{,}283 & 8.5\% & Wrong work-product type,
missing section, malformed table/redline. & Deliverable schema;
artifact-first writer; release-form gate. \\
Legal-rule use & 1{,}464 & 9.6\% & Required rule, clause,
standard, or doctrinal condition missing. & Authority retrieval;
rule-to-fact matrix; specialist skill. \\
Coverage & 935 & 6.2\% & Enumerated source, issue, clause, or
subpart left untreated. & Mandatory checklist; source and issue
closure report. \\
Other & 3{,}720 & 24.5\% & Judge reason not cleanly mapped to a named
scripted class. & Learning backlog; human taxonomy review. \\
Remedial action & 513 & 3.4\% & Risk spotted without proposed cure,
action owner, or next step. & Issue-lifecycle table; recommendation
field gate. \\
Materiality/impact & 623 & 4.1\% & Severity, priority, business
impact, or party impact absent or miscalibrated. & Materiality
rubric; severity normalization. \\
Source grounding & 610 & 4.0\% & Missing source span, exhibit,
authority, or citation link. & Citation and source-grounding audit. \\
Cross-doc synthesis & 18 & 0.1\% & Related documents or clauses
not reconciled. & Cross-document map; conflict/duplication checker. \\
\bottomrule
\end{tabularx}
\caption{Error taxonomy and corresponding framework
controls. Counts sum to $15{,}171$ failed criteria.}
\label{tab:error-taxonomy-controls}
\end{table}

\subsection{Trace defects behind the controls}

\begin{table}[H]
\centering
\scriptsize
\setlength{\tabcolsep}{3pt}
\renewcommand{\arraystretch}{1.08}
\begin{tabularx}{\textwidth}{@{}L{0.45cm}L{2.55cm}L{2.8cm}YY@{}}
\toprule
ID & Trace defect & Evidence & Legal failure & Control \\
\midrule
D1 & Long-prose edit collision & Observed in long-form drafting traces and manually reviewed regressions & Planned sections disappear during brittle rewrites & Section-aware parsing; artifact diff check \\
D2 & Referenced-not-read coverage & Observed when cited spreadsheets or tables were not opened before final answer & Spreadsheet-held entities amounts or terms are cited but never opened & Coverage manifest; mandatory spreadsheet inspection \\
D3 & Deliverable-shape mismatch & Observed in markup and redline regressions & Draft exists but final submitted work product has the wrong shape & Work-product schema; artifact-first submission \\
D4 & Silent content omission & Observed in stratified rationale and trace review & Required contracts trackers or sections are dropped & Completeness checklist before release \\
D5 & Single-turn protocol & Present by construction in the 1251-run Codex/GPT-5.5 baseline & Drafter is its own reviewer and flags cannot reopen the run & External evaluator; bounded revision pass \\
D6 & Startup skill overhead & Observed as bootstrap commands before source inspection & Cost and latency are spent before task evidence is read & Defer skill bootstrap until task files are inspected \\
D7 & No citation-grounding contract & Present by construction & Authorities need not resolve to source corpus or approved law & Citation scan; authority whitelist \\

\end{tabularx}
\caption{Qualitative trace-defect map from Codex JSONL transcripts. Rows are
mirrored in the trace-defect analysis ledger and serve as process evidence for the controls in
Table~\ref{tab:error-taxonomy-controls}.}
\label{tab:detailed-trajectory-defects}
\end{table}

\subsection{GPT-5.5 trajectory slices}

Table~\ref{tab:gpt55-trajectory-slices} gives concrete trace slices
from the same Codex/GPT-5.5 result cell used in
\S\ref{subsec:trajectory}--\ref{subsec:patterns}: observable agent
behavior, the scored miss, and the missing release gate. They make the
abstract failure classes concrete -- in each case the agent reads the full
source set yet omits a testable fact, formula, or identity check that a
deterministic gate would have caught.

\begin{table}[H]
\centering
\tiny
\setlength{\tabcolsep}{2.4pt}
\renewcommand{\arraystretch}{0.98}
\begin{tabularx}{\textwidth}{@{}L{2.35cm}L{3.15cm}YY@{}}
\toprule
Task & Trace slice & Scored error & Diagnostic \\
\midrule
HSR strategy memo & $11/11$ docs; many commands; \texttt{rg} fails then
\texttt{find}/\texttt{sed}; writes HSR and risk memos. & Omits
$\mathrm{HHI}>1{,}800$ plus $>200$ delta and PeakAir hot-document
phrases. & Hot-doc ledger; antitrust threshold gate. \\
Credit covenant extraction & $7/7$ docs; greps covenant
terms; reads agreements, certificate, and email chain; writes memo. &
Omits EBITDA formula, $15\%$/\$7.5M caps, and Sub Notes EBITDA
tightening effect. & Number ledger; cross-document calculation check. \\
Arbitral-award findings & $11/11$ docs; reads competing
award/expert files; writes award-summary memo. & Anchors to wrong
award: parties, arbitrators, and governing-law facts mismatch target
matter. & Matter-identity gate before drafting. \\
GDPR adequacy impact memo & $28/28$ docs; inventories
transfer registers, adequacy materials, DPAs, BCRs, and regulator
letters. & Analyzes different decision; misses NovaTerra, September
2026, SCC retention, Turvenia, and Article 49(1)(f). & Entity and
jurisdiction state lock. \\
Fund II LPA drafting & $8/8$ docs; writes issues memo
and LPA draft from precedent, term sheet, ESG, and investor comments. &
Missing testable clauses: $15/85$ and $80/20$ catch-up formulas and
Investment Period/recycling reconciliation. & Clause schema; formula
verifier. \\
\bottomrule
\end{tabularx}
\caption{Representative GPT-5.5 trajectory slices. Document counts are from
run metrics; misses are from paired judge rationales.}
\label{tab:gpt55-trajectory-slices}
\end{table}

\subsection{Self-evolving optimization protocol (hard-$10$)}
\label{app:optimization-protocol}

The hard-$10$ panel (Figure~\ref{fig:optimization-curve}) measures harness
learning in isolation, separately from the full-corpus results. The ten
tasks are selected for low baseline pass rate, non-trivial document load,
and enough rubric criteria to form a meaningful trajectory, with at most one
task per practice area for cross-domain coverage. One optimization
\emph{step} runs all ten tasks (solve then judge), after which the learner
proposes one batch of harness edits; the loop runs ten such steps from an
empty starting harness. Judging uses a two-rung model ladder
(a small judge, escalating to a larger one only on context overflow or
parse failure). A candidate harness is accepted only if its scored
per-task pass rate strictly exceeds the last accepted harness
(\emph{best-so-far gate}); otherwise it is rolled back and archived, which is
why the gated curve is monotone while the raw candidates are not. The learner
may edit Skills, Knowledge, and Tools under a fixed budget (at most a few
edits per step and at most one new tool file, which must be standard-library
Python under a small size limit and pass static safety checks), and it sees
only redacted signals -- never task ids, party or document names, dates,
amounts, clause numbers, or rubric text. The headline outcome (two unrelated
solver/learners converging to within $0.4$\,pp) is reported in
\S\ref{subsec:ablations}; full per-step ledgers are in the data pack.

\subsection{Reasoning-effort and document-summary ablations}
\label{app:ablation-protocols}

\paragraph{Reasoning effort.} On the same ten tasks, we sweep the
coding-agent CLI effort setting in a full factorial: \texttt{low},
\texttt{medium}, \texttt{high}, \texttt{xhigh} for Codex/GPT-5.5, and
\texttt{low}, \texttt{medium}, \texttt{high}, \texttt{xhigh}, \texttt{max}
for Claude Code/Sonnet, with \texttt{medium} as the baseline. The setting is
a CLI-level effort flag, not an API reasoning parameter, so it varies the
agent's exploration budget on a fixed harness.

\paragraph{Document summaries.} The cached document-summary tool
(\texttt{summarize\_docs}) writes a structured per-document brief -- named
entities, operative terms, quantitative and date anchors, at roughly a
quarter of the source length -- intended as an index back into the originals.
The ablation pairs the same GPT-5.4-mini Codex setting with and without the
prewarmed summary directory, on the same task set. Its conclusion
(\S\ref{subsec:ablations}) is that the cache does not close the long-input
gap; full paired outcomes are in the data pack.

\subsection{Data availability}
\label{app:data-availability}

To keep this appendix to protocols and representative cases rather than
exhaustive tables, the complete experimental record is released as an open
data pack: per-task scored outputs and judge rationales, the native solver
transcripts, the derived trace and cell metrics behind every figure, the
optimization and ablation ledgers, and the harness artifacts (Skills, Tools,
Knowledge). Readers who need a specific number not printed here can recover
it from that release. The data pack is hosted on Hugging Face at
\url{https://huggingface.co/buckets/hhhhhhejia/parthenon-harvey-results-backup},
and the $1{,}251$-skill task-routed Skills library is additionally released as a
standalone repository at \url{https://github.com/HHHHHejia/parthenon-skills}.

\endgroup

\end{document}